%% file: atlasnav_arxiv.tex
\documentclass{article}
\usepackage{preprint_layout,times}

\input{math_commands.tex}

\usepackage{graphicx}
\usepackage{booktabs}
\usepackage{float}
\usepackage{placeins}
\usepackage{hyperref}
\usepackage{url}
\usepackage{fontawesome5}
\hypersetup{
  hidelinks,
  pdftitle={AtlasNav: Mitigating Evidence Blindness with Persistent Corpus Navigation},
  pdfauthor={Hongyu Guo, Zhiyu Zheng, Minghao Hu, Shuai Lei, Zhunchen Luo, Chao Ma, Yunbo Cao, Zhao Cao}
}
\renewcommand{\headrulewidth}{0pt}

\graphicspath{{figures/}}

\input{paper_macros.tex}

\title{AtlasNav: Mitigating Evidence Blindness\\
with Persistent Corpus Navigation\par\vspace*{-11.85pt}}

\author{
  \bfseries Hongyu Guo$^{\dagger}$,
  Zhiyu Zheng$^{\dagger}$,
  Minghao Hu,
  Shuai Lei,
  Zhunchen Luo,\\[2pt]
  \textbf{Chao Ma},
  \textbf{Yunbo Cao},
  \textbf{Zhao Cao}$^{*}$\\[4pt]
  \mdseries\normalsize
  Gaoling School of Artificial Intelligence, Renmin University of China\\
  Center of Information Research, AMS, Beijing, China\\
  Center for Large Model Intelligence Application, Beijing, China.\\[6pt]
  \mdseries\normalsize
  \href{https://github.com/fresnellll/AtlasNav}{\faGithub\ AtlasNav}%
  \hspace{2em}%
  \href{https://huggingface.co/datasets/fresnellll/atlasnav-artifacts}{%
    \raisebox{-0.19ex}{\includegraphics[height=2.09ex]{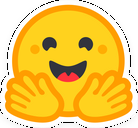}}\ AtlasNav}
}

\iclrfinalcopy

\begin{document}

\maketitle

\input{sections/00_abstract}
\input{sections/01_introduction}
\input{sections/02_related_work}
\begingroup
\setlength{\textfloatsep}{6pt plus 1pt minus 1pt}
\input{sections/03_method}
\input{sections/04_experiments}
\endgroup
\input{sections/05_conclusion}

\input{sections/06_statements}

\bibliography{references}
\bibliographystyle{preprint_references}

\appendix
\input{sections/appendix}

\end{document}

%% file: math_commands.tex
\usepackage{amsmath,amsfonts,bm}

\def\eqref#1{equation~\ref{#1}}

\def\1{\bm{1}}

\DeclareMathAlphabet{\mathsfit}{\encodingdefault}{\sfdefault}{m}{sl}
\SetMathAlphabet{\mathsfit}{bold}{\encodingdefault}{\sfdefault}{bx}{n}



%% file: paper_macros.tex
\newcommand{\runinlabel}[1]{\noindent\textbf{#1}\ }

%% file: sections/00_abstract.tex
\begin{abstract}
As language-model agents become more capable of iterative search, corpus access is shifting from retrieval toward interaction. Agents can explore the corpus, inspect documents, and use newly discovered evidence to decide what to examine next. Yet accessible evidence may still fail to become usable within a finite interaction budget. We call this progressive failure \emph{Evidence Blindness}: supporting documents may never enter view, may remain unopened, or may fail to expose the decisive evidence even after being opened. A key reason is that agents often have to infer useful evidence directions during interaction, spending limited budget on deciding where to search next. Existing approaches either leave corpus structure largely implicit or reconstruct useful directions at query time. AtlasNav instead organizes reusable cross-document structure before any query arrives. It builds a persistent multi-view Corpus Atlas, which each query can navigate adaptively while still accessing the original documents directly. On BrowseComp-Plus, AtlasNav outperforms the previous state-of-the-art interactive corpus access method across different backbones. On DeepSeek, it improves strict accuracy by 7.47 points while reducing query-time inference cost by 30.22\%. AtlasNav also reduces Evidence Blindness, realizes complete evidence earlier, remains robust to corpus-structure and scale shifts on PhantomWiki, and achieves leading performance on heterogeneous enterprise data.
\end{abstract}

%% file: sections/01_introduction.tex
\section{Introduction}
\label{sec:introduction}

Traditional retrieval-augmented generation (RAG) accesses external knowledge through retrieval, selecting query-relevant content for downstream reasoning \citep{lewis2020retrieval}. As language-model agents become increasingly capable, corpus access is shifting from retrieval toward interaction: agents directly interact with the corpus, iteratively exploring it and using newly discovered evidence to decide what to examine next \citep{li2026dci,lu2026drdci,zhuang2026rise}. This iterative interaction keeps the corpus broadly accessible while allowing evidence discovery to adapt continuously as new information emerges. We refer to this emerging paradigm as \emph{interactive corpus access}. Prior work has shown that, by allowing agents to access and explore the corpus more directly and flexibly, \emph{interactive corpus access} can unlock substantially stronger agentic search performance than conventional retrieval-based access on challenging tasks \citep{li2026dci,lu2026drdci,zhuang2026rise}.

Despite its promise, existing work on \emph{interactive corpus access} has yet to fully exploit and organize the latent structure of the corpus. Existing approaches either leave such structure largely implicit in the raw corpus or reconstruct only a query-specific portion of it during interaction \citep{li2026dci,lu2026drdci,zhuang2026rise}. Without reusable corpus organization, agents must infer useful evidence directions online, making it difficult to discover the right evidence early under a finite interaction budget. This leads to both lower search accuracy, as relevant evidence is more easily missed, and higher inference cost, as agents repeatedly explore the corpus to determine where to look next. These outcome-level failures point to a deeper process-level question: how does evidence that remains accessible in the corpus fail to become usable as interaction unfolds? Existing work has not yet examined this process systematically.

To understand this process, we identify a systematic failure mode of \emph{interactive corpus access}, which we call \emph{Evidence Blindness}: evidence may remain available in the corpus yet fail to become usable within a finite interaction budget, leaving the agent to keep searching or reason from an incomplete evidence state until the budget is exhausted. We diagnose this silent failure progressively, from coarse corpus access to fine-grained evidence realization. Required evidence may exist in the accessible corpus but never enter the agent's view; a supporting document may come into view but never be opened; and even an opened document may fail to expose the decisive evidence. This progressive diagnosis shows exactly where evidence stops becoming usable, down to whether the decisive evidence is ever reached.

The diagnosis above suggests that improving \emph{interactive corpus access} requires more than better online search: the corpus itself must provide reusable structure that helps evidence become visible and usable within a finite interaction budget. AtlasNav follows this principle by moving corpus organization outside the query-time loop. It builds a persistent navigation structure that captures complementary cross-document connections, consistent with evidence that multi-hop reasoning can exhibit multi-view structure \citep{liu2026parallax}. This structure is reused across queries, allowing online interaction to focus on selecting and following promising directions rather than repeatedly reconstructing the search space. By shifting interaction from repeated rediscovery toward guided navigation, AtlasNav converts more of the finite budget into evidence realization. Useful evidence can therefore become available earlier and more completely, reducing Evidence Blindness and enabling stronger task performance with less inference cost.

\textbf{Our contributions are threefold:}
\begin{itemize}
    \item We identify Evidence Blindness as a systematic failure mode of \emph{interactive corpus access}. We characterize how accessible evidence can progressively fail to become usable under finite interaction budgets, and diagnose this failure from coarse corpus access to fine-grained evidence realization.
    \item We introduce AtlasNav, a persistent corpus-navigation framework for \emph{interactive corpus access}. By moving reusable corpus organization outside the query-time loop and exposing complementary cross-document connections through a shared navigation structure, AtlasNav replaces repeated query-time reconstruction with query-adaptive navigation.
    \item We show that persistent corpus navigation improves both evidence realization and search efficiency. On BrowseComp-Plus \citep{chen2025browsecompplus}, AtlasNav consistently outperforms DR-DCI \citep{lu2026drdci}, the previous state-of-the-art \emph{interactive corpus access} method, across four backbones; on DeepSeek, it improves accuracy by 7.47 points while reducing inference cost by 30.22\%, alongside lower Evidence Blindness and faster convergence to the evidence-supplied reference. The same design remains effective on million-scale PhantomWiki and heterogeneous enterprise data.
\end{itemize}

%% file: sections/02_related_work.tex
\section{Related Work}
\label{sec:related-work}

\runinlabel{Corpus access for agentic search.}
RAG and agentic RAG expose external knowledge primarily through retrieval, including iterative retrieval and query reformulation \citep{lewis2020retrieval,trivedi2023ircot,jin2025searchr1,li2025webthinker}. Direct Corpus Interaction (DCI) directly exposes the corpus to the agent, while DR-DCI and RISE construct query-conditioned interaction spaces online for scalable agentic search \citep{li2026dci,lu2026drdci,zhuang2026rise}. AtlasNav instead reuses persistent cross-document organization across queries.

\runinlabel{Structured corpus representations.}
Hierarchical, graph-based, and multi-view methods organize corpus structure for retrieval and multi-hop reasoning, including RAPTOR, GraphRAG, HippoRAG, SiReRAG, and Think Parallax \citep{sarthi2024raptor,edge2024graphrag,gutierrez2024hipporag,zhang2025sirerag,liu2026parallax}. AtlasNav uses such structure as a persistent navigation interface rather than primarily for retrieval.

\runinlabel{Evidence-access diagnostics.}
RAGAS, ARES, and RAGChecker diagnose retrieval and generation beyond final accuracy \citep{es2024ragas,saadfalcon2024ares,ru2024ragchecker}, while DCI measures trajectory-level Coverage and Localization \citep{li2026dci}. Trust No Tool studies trajectory-level failures induced by untrusted tool feedback \citep{yan2026trusttool}. Evidence Blindness instead localizes where benchmark-grounded evidence fails to become usable during corpus interaction.

%% file: sections/03_method.tex
\begin{figure}[!t]
  \centering
  \includegraphics[width=\linewidth]{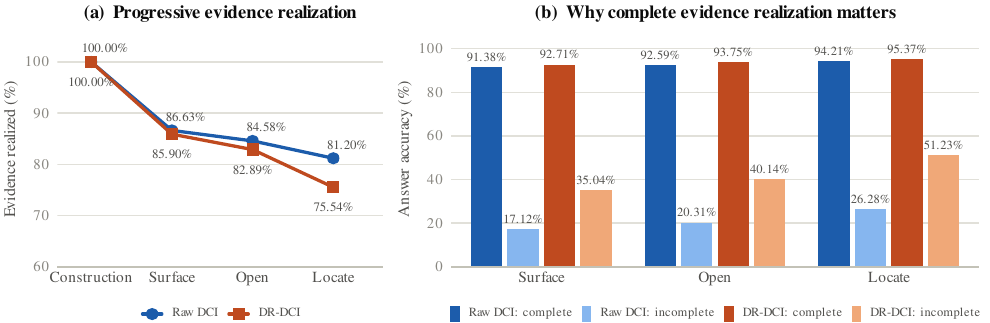}
  \caption{Evidence realization on BrowseComp-Plus with DeepSeek-V4-Flash.
  (a) Complete evidence realized per checkpoint. (b) Strict accuracy given
  complete vs.\ incomplete realization.}
  \label{fig:evidence_realization}
  \vspace*{10pt}
\end{figure}

\section{Evidence Blindness}
\label{sec:evidence_blindness}

\subsection{Evidence Blindness in Interactive Corpus Access}
\label{sec:evidence_blindness_observation}

\begin{figure}[!t]
  \centering
  \includegraphics[width=\linewidth]{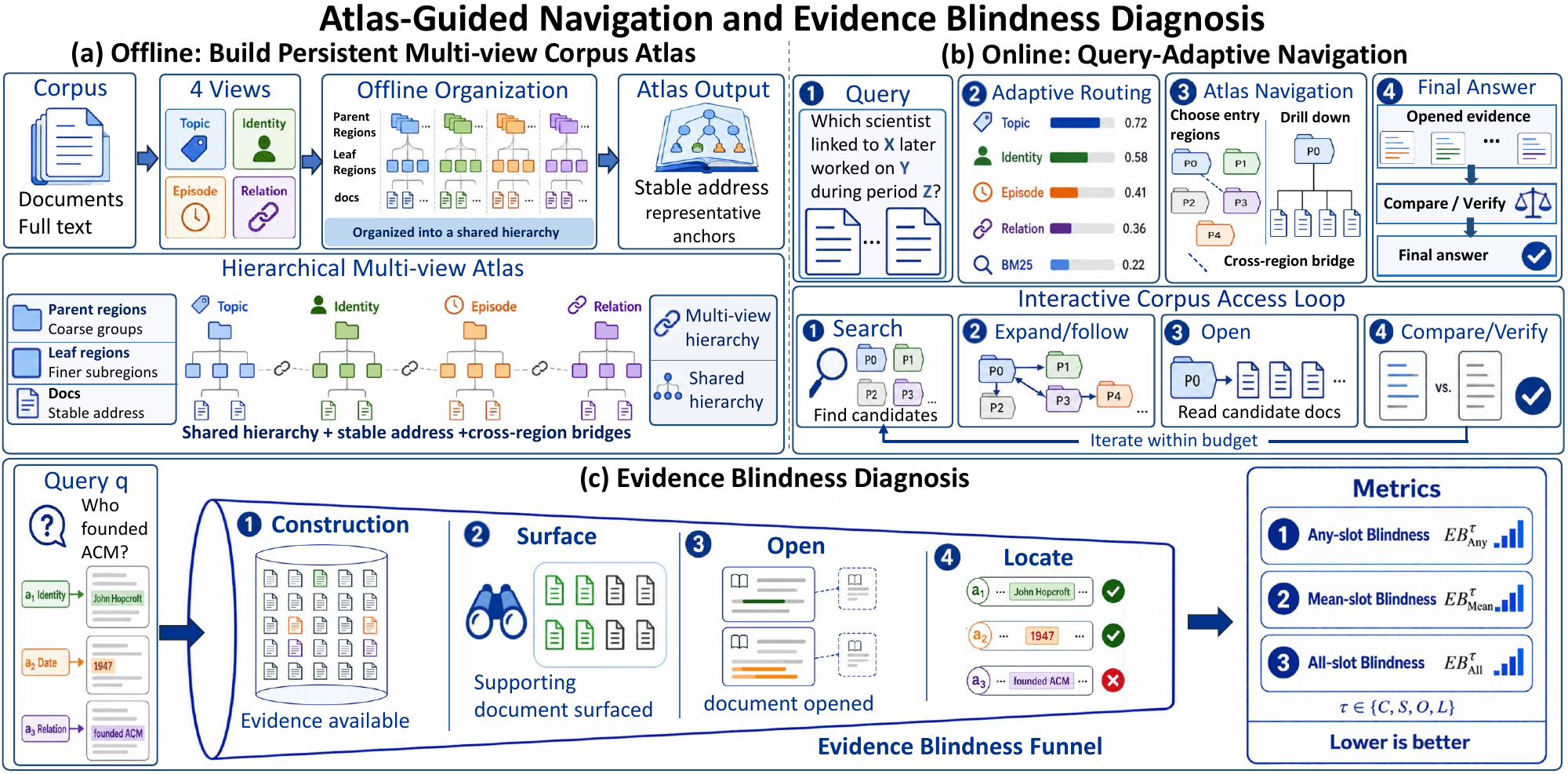}
  \caption{AtlasNav and Evidence Blindness. (a) Offline construction of the persistent multi-view Corpus Atlas. (b) Query-adaptive finite-budget navigation. (c) Evidence Blindness diagnosis across the four evidence-realization checkpoints.}
  \label{fig:atlasnav_overview}
  \label{fig:evidence_blindness}
  \vspace{6pt}
\end{figure}

\emph{Interactive corpus access} unfolds through a repeated interaction loop. The agent searches for candidate evidence, follows newly discovered directions, opens promising documents, and compares or verifies the observed information before deciding whether to continue. Figure~\ref{fig:atlasnav_overview}(b) illustrates this Interactive Corpus Access Loop. Evidence realization can be tracked along the same trajectory: before interaction, the required evidence must first be available in the accessible corpus; during search and exploration, supporting documents may enter the agent's view; opened documents then expose their contents; and continued inspection may finally reveal the decisive evidence needed for answering. Figure~\ref{fig:atlasnav_overview}(c) summarizes these progressively finer evidence-realization checkpoints as Construction, Surface, Open, and Locate; notation is summarized in Appendix~\ref{app:notation_summary}.

To make this failure directly observable, we use BrowseComp-Plus, where all annotated supporting evidence is present in the accessible corpus and the model must gather, integrate, and reason over that evidence to answer correctly. We examine two representative \emph{interactive corpus access} interfaces: DCI, which exposes the corpus directly to the agent, and DR-DCI, which organizes access through a query-conditioned dynamic workspace \citep{li2026dci,lu2026drdci}. Figure~\ref{fig:evidence_realization} tracks how much of the required evidence remains realized under these interfaces as interaction proceeds. Panel~(a) shows a progressive drop from Construction through Surface, Open, and Locate, even though the corpus itself already contains all required evidence. Panel~(b) shows why this loss matters: when the complete evidence set is realized, answer accuracy remains above 90\%, whereas incomplete realization leads to a substantial accuracy drop at every checkpoint. We call this progressive failure to turn accessible evidence into usable evidence \emph{Evidence Blindness}.

\subsection{Formalizing Evidence Blindness}
\label{sec:evidence_blindness_formalization}

To formalize Evidence Blindness, we track the required evidence for each question across the four checkpoints introduced above. For a question $q$, let $\mathcal{E}_q=\{e_1,\ldots,e_{m_q}\}$ denote its required \emph{evidence slots}, where each slot corresponds to a factual constraint needed to answer the question. Let $\tau\in\mathcal{T}=\{C,S,O,L\}$ index Construction, Surface, Open, and Locate. Operationally, Construction requires the evidence to exist in the accessible corpus; Surface requires a supporting document to appear in an agent-visible observation; Open requires the supporting document's original content to enter model-visible context; and Locate requires the decisive evidence itself to enter that context. Let $\mathcal{E}_q^\tau\subseteq\mathcal{E}_q$ denote the evidence slots realized by checkpoint $\tau$.

Because a slot cannot be realized at a finer checkpoint without satisfying the preceding ones, evidence realization forms a monotonic funnel:
\begin{equation}
  \mathcal{E}_q^C \supseteq \mathcal{E}_q^S \supseteq \mathcal{E}_q^O \supseteq \mathcal{E}_q^L.
  \label{eq:evidence_funnel}
\end{equation}
To quantify how much required evidence survives to each checkpoint, we define the realization ratio $r_q^\tau=|\mathcal{E}_q^\tau|/|\mathcal{E}_q|$. Thus, $r_q^\tau=1$ means that all required evidence slots have been realized by checkpoint $\tau$, while smaller values indicate partial evidence loss.

To distinguish different degrees of evidence loss across a question set $\mathcal{Q}$, we define three complementary Evidence Blindness metrics:
\begin{equation}
  \operatorname{EB}_{\mathrm{Any}}^{\tau}
    = \frac{1}{|\mathcal{Q}|}\sum_{q\in \mathcal{Q}}\mathbf{1}[r_q^\tau=0],
  \quad \operatorname{EB}_{\mathrm{Mean}}^{\tau}
    = 1-\frac{1}{|\mathcal{Q}|}\sum_{q\in \mathcal{Q}}r_q^\tau,
  \quad
  \operatorname{EB}_{\mathrm{All}}^{\tau}
    = \frac{1}{|\mathcal{Q}|}\sum_{q\in \mathcal{Q}}\mathbf{1}[r_q^\tau<1].
  \label{eq:evidence_blindness}
\end{equation}
Here, $\operatorname{EB}_{\mathrm{Any}}^{\tau}$ measures the fraction of questions for which no required evidence is realized, $\operatorname{EB}_{\mathrm{Mean}}^{\tau}$ measures the average fraction of missing evidence, and $\operatorname{EB}_{\mathrm{All}}^{\tau}$ measures the fraction for which the complete required evidence set is not realized. Together, they distinguish total failure, average evidence loss, and incomplete evidence realization.

To make the four checkpoints observable in practice, we derive them from the interaction trajectory. Construction is saturated in our main setting because all annotated supporting evidence is present in the accessible corpus. Surface and Open are directly determined from document-level trajectory events, while Locate additionally requires checking whether the decisive evidence itself entered model-visible context. We therefore use a fragment-level reference fixed before evaluation that aligns each evidence slot with acceptable decision-relevant fragments; construction and matching details are provided in Appendix~\ref{app:locate_construct}. This operationalization lets Evidence Blindness localize where usable evidence is lost, which final-answer accuracy alone cannot reveal.

\section{AtlasNav: Persistent Corpus Navigation}
\label{sec:method}
\label{sec:atlasnav}

Evidence Blindness reveals that the main difficulty is not whether evidence exists in the corpus, but whether finite interaction can turn it into usable evidence before the budget is exhausted. AtlasNav addresses this bottleneck by organizing reusable cross-document structure in advance, so query-time interaction can focus on navigating promising evidence directions rather than reconstructing them from scratch.

\subsection{Finite-Budget Navigation}
\label{sec:finite_budget_navigation}

We evaluate corpus interfaces by how efficiently a fixed agent turns finite interaction into task performance. Let $\mathcal{C}$ denote the original accessible corpus, $M$ the language-model agent, $I$ a corpus interface, and $B$ the interaction budget. For a question $q$, $M$ interacts with $\mathcal{C}$ through $I$ and answers within $B$; the agent and budget protocol are fixed across interfaces. Let $A_I(B)$ denote strict task performance under budget $B$; notation is summarized in Appendix~\ref{app:notation_summary}.

Endpoint performance alone does not reveal how efficiently that performance is reached. We therefore use the same agent supplied directly with benchmark-designated gold documents as an empirical reference, with performance $A_{\mathrm{ref}}$, and define
\begingroup
\setlength{\abovedisplayskip}{2pt}
\setlength{\belowdisplayskip}{2pt}
\setlength{\abovedisplayshortskip}{2pt}
\setlength{\belowdisplayshortskip}{2pt}
\begin{equation}
  \Delta_I(B)=A_{\mathrm{ref}}-A_I(B).
  \label{eq:reference_gap}
\end{equation}
\endgroup
A smaller or faster-closing $\Delta_I(B)$ means that the interface realizes more of the agent's evidence-conditioned capability within the available budget. $A_{\mathrm{ref}}$ is an empirical reference, not a theoretical upper bound. Together, Evidence Blindness and $\Delta_I(B)$ characterize process-level evidence loss and its task-level consequence, respectively.

\subsection{Building a Persistent Multi-View Corpus Atlas}
\label{sec:atlasnav_representation}

To reduce Evidence Blindness under a finite interaction budget, AtlasNav must prevent early observations from over-concentrating exploration along a single semantic direction. Interactive search is inherently history-conditioned, with each action depending on the trajectory observed so far \citep{gandhi2025gobrowse}. Yet useful cross-document evidence may be connected not only by topic, but also by shared referents, event progression, and explicit relations \citep{das2019multistep,lu2023multihop,yang2025eventrag,liu2026parallax}. AtlasNav therefore uses four views: Topic for thematic orientation, Identity for referential continuity, Episode for events, times, places, and stages, and Relation for roles, causes, comparisons, and other cross-item relations. Together, these views broaden the evidence directions available during interaction, helping the agent surface complementary evidence earlier within the same budget.

To turn the four forms of continuity into reusable navigation structure, let $\mathcal{D}$ denote the set of original documents and $\mathcal{V}=\{\mathrm{Top},\mathrm{Id},\mathrm{Epi},\mathrm{Rel}\}$ the four views: Topic, Identity, Episode, and Relation. AtlasNav extracts a view-specific signature $\operatorname{Sig}_v(d)$ from each document $d\in\mathcal{D}$ for every view $v\in\mathcal{V}$. A shared encoder maps this signature into a view-specific representation, from which AtlasNav constructs a sparse neighborhood graph:
\begin{equation}
  z_d^v=\operatorname{Enc}\!\left(\operatorname{Sig}_v(d)\right),
  \qquad G^v=(\mathcal{D},E^v,W^v),
  \qquad d\in\mathcal{D},\; v\in\mathcal{V}.
  \label{eq:multiview_representation}
\end{equation}
Here, $E^v$ connects nearby documents according to their representations $z_d^v$, and $W^v$ records the corresponding affinities. To preserve useful routes across regions with different semantic densities, AtlasNav normalizes neighbor distances by local density and retains weak one-way links for rare patterns. This yields complementary navigation neighborhoods, giving the agent alternative cross-document directions when one view is insufficient. The resulting topologies are substantially different across views: they share only about 26--39\% of their edges, while nearly 79\% of documents acquire a different strongest neighbor in at least one view. Thus, the same document can lead the agent toward different candidates depending on the evidence relation being followed.

To control the branching space exposed to the agent, AtlasNav organizes the four view-specific graphs into a role-aware conditional hierarchy. Topic and Identity define coarse parent regions, while Episode and Relation refine each parent into finer leaf regions:
\begin{equation}
  \mathcal{P}=\operatorname{Cluster}_{\mathrm{multi}}(G^{\mathrm{Top}},G^{\mathrm{Id}}),
  \qquad
  \mathcal{L}(P)=\operatorname{Cluster}_{\mathrm{multi}}\!\left(G^{\mathrm{Epi}}[P],G^{\mathrm{Rel}}[P]\right),
  \quad P\in\mathcal{P}.
  \label{eq:conditional_hierarchy}
\end{equation}
Here, $\mathcal{P}$ is the set of parent regions, $\mathcal{L}(P)$ is the set of leaf regions nested within parent $P$, $G^v[P]$ is view $v$'s subgraph restricted to $P$, and $\operatorname{Cluster}_{\mathrm{multi}}$ jointly partitions the view-specific graphs using Leiden optimization \citep{mucha2010community,traag2019leiden}. The resulting hierarchy reduces branching while preserving finer evidence transitions within each region.

To expose the hierarchy as a navigation interface, AtlasNav assigns each document a stable parent--leaf address, summarizes regions with representative anchors, and preserves strong cross-region transitions through Episode- and Relation-level bridges. These structures support coarse-to-fine navigation without pruning access to the original corpus. Detailed construction of addresses, anchors, bridges, and hierarchy parameters is provided in Appendices~\ref{app:atlas_signatures}--\ref{app:atlas_access_points}. Together, these components form a persistent navigation space that preserves complementary cross-document structure for query-adaptive exploration.

\subsection{Query-Adaptive Atlas Navigation}
\label{sec:atlasnav_navigation}

To adapt the persistent Atlas to each query, AtlasNav routes interaction through five complementary channels. Four channels---Topic, Identity, Episode, and Relation---correspond to the semantic views constructed in Section~\ref{sec:atlasnav_representation}. A fifth BM25 channel preserves exact lexical signals that semantic representations may smooth over, such as rare names, phrases, or identifiers \citep{robertson2009bm25}. To decide which directions deserve more attention for the current query, AtlasNav uses a query-adaptive router that assigns a weight to each channel. These weights shift the limited interaction budget toward the most relevant evidence directions while retaining complementary signals from the others.

To combine the five weighted channels into a common document priority, AtlasNav uses weighted Reciprocal Rank Fusion (RRF) \citep{cormack2009rrf}. For a query $q$, each channel first scores the documents according to its own matching signal and produces an independent ranking. Let $\operatorname{rank}_c(d\mid q)$ denote the position of document $d$ in the ranking produced by channel $c$, and let $w_c(q)$ be the weight assigned to that channel by the router. Because the four semantic views and BM25 produce scores on different scales, AtlasNav fuses their rankings rather than calibrating the raw scores directly:
\begingroup
\setlength{\abovedisplayskip}{2pt}
\setlength{\belowdisplayskip}{2pt}
\setlength{\abovedisplayshortskip}{2pt}
\setlength{\belowdisplayshortskip}{2pt}
\begin{equation}
  s_q(d)=\sum_{c\in\mathcal{V}\cup\{\mathrm{BM25}\}}
  \frac{w_c(q)}{\kappa+\operatorname{rank}_c(d\mid q)}.
  \label{eq:weighted_rrf}
\end{equation}
\endgroup
Here, $\kappa$ is the RRF smoothing constant. This fusion gives more influence to channels that are more relevant to the current query, while still allowing highly ranked evidence from complementary channels to contribute. The resulting $s_q(d)$ provides a single query-adaptive priority over documents.

To turn document priorities into actionable navigation, AtlasNav projects the ranked documents back onto the persistent Atlas regions. Regions supported by high-priority documents become initial entry points, and the agent then navigates from these regions toward finer leaves, neighboring regions, and original documents. This lets limited observations cover several promising parts of the evidence space early, rather than spending the budget on repeatedly rediscovering where to search. From these entries, the downstream agent continues its normal interaction loop---searching, opening, comparing, and verifying evidence---while retaining direct access to the original corpus. The router is trained only on corpus-derived single- and multi-document tasks; BrowseComp-Plus evaluation questions and labels are excluded, with training details in Appendix~\ref{app:router_tasks}. Together, query-adaptive routing and Atlas navigation turn persistent corpus structure into query-specific evidence paths without reconstructing the search space online.

%% file: sections/04_experiments.tex
\section{Experiments}
\label{sec:experiments}
We evaluate AtlasNav across open-domain agentic search, controlled corpus shifts, and heterogeneous enterprise knowledge, focusing on evidence realization and finite-budget efficiency. Additional diagnostics and ablations are provided in Appendices~\ref{app:complete_results}--\ref{app:auxiliary}.

\subsection{Experimental Setup}
\label{sec:experimental_setup}
\runinlabel{Benchmarks.}
We evaluate on three complementary settings. BrowseComp-Plus \citep{chen2025browsecompplus} is the main benchmark for large-scale open-domain agentic search. PhantomWiki \citep{gong2025phantomwiki} tests controlled structural and scale transfer by fixing 200 questions and their supports while expanding the corpus from 10,059 to 1,006,839 files. EnterpriseRAG-Bench \citep{sun2026enterpriseragbench} tests transfer to heterogeneous enterprise knowledge with 511,958 documents across nine source types.

\runinlabel{Models.}
Corpus and query embeddings use Qwen3.7-Text-Embedding \citep{alibabacloud2026qwen37embedding} throughout. BrowseComp-Plus uses four backbones---DeepSeek-V4-Flash \citep{deepseekai2026v4}, MiMo-V2.5 \citep{xiaomi2026mimov25}, ChatGPT-5.6-Luna \citep{openai2026gpt56}, and Qwen-3.7-Flash \citep{alibabacloud2026qwen37flash}---to test robustness across model families and capability--cost regimes. ChatGPT provides a strong proprietary backbone, DeepSeek a high-capability cost-efficient setting, and MiMo and Qwen broaden model diversity. PhantomWiki and EnterpriseRAG-Bench use DeepSeek-V4-Flash throughout.

\runinlabel{Baselines.}
We compare DCI, which directly interacts with the full corpus \citep{li2026dci}; DR-DCI, which builds a query-conditioned workspace online \citep{lu2026drdci}; and AtlasNav, which navigates a persistent Corpus Atlas. BrowseComp-Plus additionally includes the same-agent evidence-supplied empirical reference from Section~\ref{sec:finite_budget_navigation}.

\runinlabel{Metrics.}
We report benchmark-defined strict accuracy, query-time agent inference cost, Evidence Blindness at the Surface, Open, and Locate checkpoints defined in Section~\ref{sec:evidence_blindness_formalization}, and the empirical reference gap $\Delta_I(B)$ from Section~\ref{sec:finite_budget_navigation}. Costs are compared only within backbone. EnterpriseRAG-Bench follows its official Overall, Correctness, Completeness, Document Recall, and Invalid Extra metrics \citep{enterpriseragleaderboard2026}; full definitions are in Appendix~\ref{app:experimental_details}. Full protocols and exact values are in Appendices~\ref{app:complete_main_results}--\ref{app:experimental_details}.

\subsection{Evidence Realization and Finite-Budget Efficiency}
\label{sec:browsecomp_results}
We examine the endpoint accuracy--efficiency frontier, localize the improvement with Evidence Blindness, and trace turn- and cost-budget trajectories to test whether it emerges early.

\runinlabel{Accuracy--efficiency.}
Table~\ref{tab:browsecomp_main} shows that AtlasNav improves the accuracy--efficiency frontier across all four backbones. Relative to DR-DCI, it raises strict accuracy by 3.98--21.57 percentage points while reducing query-time agent inference cost by 0.62--30.22\%. The gains are larger on backbones that are less effective at locating and assembling evidence through search alone, a capability directly stressed by BrowseComp-style agentic search evaluations \citep{wei2025browsecomp}. This suggests that persistent navigation is particularly valuable when the backbone itself provides weaker search guidance.

\begingroup
\setlength{\intextsep}{7pt plus 1pt minus 1pt}
\setlength{\abovecaptionskip}{0pt}
\setlength{\belowcaptionskip}{4pt}
\begin{table}[H]
  \caption{Accuracy and recorded query-time agent inference cost on BrowseComp-Plus. Costs are normalized within each backbone to AtlasNav $=1.000$; accuracy gains and cost savings are relative to DR-DCI.}
  \label{tab:browsecomp_main}
  \centering
  \small
  \setlength{\tabcolsep}{3.5pt}
  \renewcommand{\arraystretch}{1.08}
  \begin{tabular*}{\textwidth}{@{\extracolsep{\fill}}l|ccccc|cccc@{}}
    \toprule
    & \multicolumn{5}{c|}{Strict Accuracy (\%) $\uparrow$}
    & \multicolumn{4}{c}{Normalized Query-Time Agent Cost $\downarrow$} \\
    \cmidrule(lr){2-6}\cmidrule(l){7-10}
    Backbone & DCI & DR-DCI & AtlasNav & Ref. & $\Delta$ Acc.
    & DCI & DR-DCI & AtlasNav & Cost save (\%) \\
    \midrule
    GPT-5.6  & 87.35 & 87.83 & \textbf{91.81} & 96.51 & $+3.98$
             & 1.718 & 1.173 & \textbf{1.000} & \textbf{14.74} \\
    DeepSeek & 81.45 & 84.58 & \textbf{92.05} & 96.51 & $+7.47$
             & 2.347 & 1.433 & \textbf{1.000} & \textbf{30.22} \\
    Qwen     & 36.99 & 50.96 & \textbf{72.53} & 95.42 & $+21.57$
             & 2.243 & 1.271 & \textbf{1.000} & \textbf{21.31} \\
    MiMo     & 62.89 & 71.08 & \textbf{79.04} & 96.14 & $+7.96$
             & 2.579 & 1.006 & \textbf{1.000} & \textbf{0.62} \\
    \bottomrule
  \end{tabular*}
\end{table}
\endgroup

\runinlabel{Evidence Blindness.}
Construction is saturated for all three interfaces and is therefore omitted from Table~\ref{tab:browsecomp_eb}.

\begingroup
\setlength{\intextsep}{7pt plus 1pt minus 1pt}
\setlength{\abovecaptionskip}{0pt}
\setlength{\belowcaptionskip}{4pt}
\begin{table}[H]
  \caption{Evidence Blindness (EB) on BrowseComp-Plus (\%); lower is better.}
  \label{tab:browsecomp_eb}
  \centering
  \small
  \renewcommand{\arraystretch}{1.06}
  \begin{tabular*}{\textwidth}{@{\extracolsep{\fill}}ll|rr|rr|rr@{}}
    \toprule
    & & \multicolumn{2}{c|}{Surface $\downarrow$}
        & \multicolumn{2}{c|}{Open $\downarrow$}
        & \multicolumn{2}{c}{Locate $\downarrow$} \\
    \cmidrule(lr){3-4}\cmidrule(lr){5-6}\cmidrule(l){7-8}
    Backbone & Interface
        & $\mathrm{EB}_{\mathrm{Any}}^{S}$ & $\mathrm{EB}_{\mathrm{All}}^{S}$
        & $\mathrm{EB}_{\mathrm{Any}}^{O}$ & $\mathrm{EB}_{\mathrm{All}}^{O}$
        & $\mathrm{EB}_{\mathrm{Any}}^{L}$ & $\mathrm{EB}_{\mathrm{All}}^{L}$ \\
    \midrule
    GPT-5.6 & DCI      & 6.63 & 6.99 & 12.89 & 13.01 & 15.90 & 16.27 \\
            & DR-DCI   & 9.40 & 9.76 & 10.48 & 11.08 & 15.90 & 16.63 \\
            & \textbf{AtlasNav} & \textbf{5.66} & \textbf{6.02} & \textbf{9.04} & \textbf{9.52} & \textbf{11.93} & \textbf{12.65} \\
    \addlinespace[2pt]
    DeepSeek & DCI      & 13.13 & 13.37 & 14.70 & 15.42 & 17.83 & 18.80 \\
             & DR-DCI   & 13.49 & 14.10 & 16.51 & 17.11 & 23.61 & 24.46 \\
             & \textbf{AtlasNav} & \textbf{4.82} & \textbf{4.94} & \textbf{7.59} & \textbf{7.83} & \textbf{10.72} & \textbf{11.45} \\
    \addlinespace[2pt]
    Qwen & DCI      & 54.70 & 55.54 & 57.35 & 58.80 & 61.57 & 63.49 \\
         & DR-DCI   & 49.04 & 49.76 & 50.60 & 51.33 & 56.63 & 57.23 \\
         & \textbf{AtlasNav} & \textbf{19.64} & \textbf{20.12} & \textbf{22.29} & \textbf{23.01} & \textbf{29.40} & \textbf{30.36} \\
    \addlinespace[2pt]
    MiMo & DCI      & 29.52 & 29.52 & 33.86 & 33.86 & 36.87 & 36.87 \\
         & DR-DCI   & 25.90 & 26.27 & 26.99 & 27.59 & 31.69 & 32.29 \\
         & \textbf{AtlasNav} & \textbf{16.75} & \textbf{17.83} & \textbf{19.88} & \textbf{20.24} & \textbf{23.73} & \textbf{24.22} \\
    \bottomrule
  \end{tabular*}
\end{table}
\endgroup

Table~\ref{tab:browsecomp_eb} shows that AtlasNav reduces evidence loss from the earliest observable interaction checkpoint. At Surface, $\operatorname{EB}_{\mathrm{All}}^{S}$ drops from 14.10\% to 4.94\% on DeepSeek and from 49.76\% to 20.12\% on Qwen relative to DR-DCI, with the same direction across all four backbones. This is where the persistent multi-view Atlas is expected to matter first: by exposing complementary evidence directions before interaction, it increases the chance that required evidence enters the agent's view early. The advantage then persists through Open and Locate; on DeepSeek, $\operatorname{EB}_{\mathrm{All}}^{S/O/L}$ decreases from 14.10/17.11/24.46\% to 4.94/7.83/11.45\% across Surface/Open/Locate. Thus, improved evidence visibility is not lost downstream, but continues to translate into opened and ultimately usable evidence.

\runinlabel{Finite-budget convergence.}
Figure~\ref{fig:budget_convergence} tests whether the reusable evidence directions built in Section~\ref{sec:atlasnav_representation} become useful early enough to matter under limited interaction. AtlasNav reaches a lower empirical reference gap $\Delta_I(B)$ sooner under both turn and cost budgets for GPT-5.6 and DeepSeek; the same trend holds for Qwen and MiMo in Appendix~\ref{app:complete_main_results}. This pattern is consistent with the role of the persistent Atlas: because candidate evidence directions are organized before query time, early interaction can begin from structured navigation points rather than spending the same budget reconstructing where to search.

The matched-budget Evidence Blindness results further show that this faster convergence comes from earlier evidence realization. At the lowest matched DeepSeek cost budget (CNY~0.15), AtlasNav reduces $\operatorname{EB}_{\mathrm{All}}^{L}$ to 35.90\%, compared with 45.18\% for DR-DCI, while the corresponding reference gap falls from 46.15 to 25.55 points. On GPT-5.6 at 60 turns, $\operatorname{EB}_{\mathrm{All}}^{L}$ is 13.61\% versus 20.72\%, with reference gaps of 7.47 and 15.55 points. By prioritizing query-relevant Atlas directions through adaptive routing and weighted fusion, AtlasNav turns more of the available interaction budget into opened and localized evidence.

\begingroup
\begin{figure}[!t]
  \centering
  \includegraphics[width=\linewidth,trim=0 5pt 0 3pt,clip]{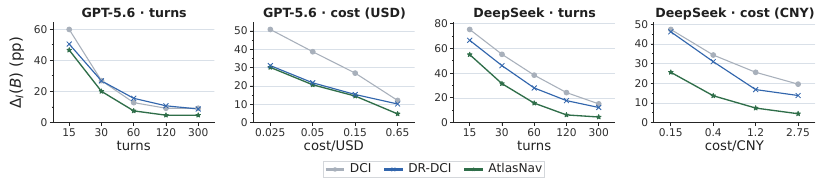}
  \vspace{-10pt}
  \setlength{\abovecaptionskip}{2pt}
  \caption{Finite-budget empirical reference gaps $\Delta_I(B)$ under turn and query-time inference-cost budgets for GPT-5.6 and DeepSeek; lower is better. Results for Qwen and MiMo are provided in Appendix~\ref{app:complete_main_results}.}
  \label{fig:budget_convergence}
  \vspace{4pt}
\end{figure}
\endgroup

Together, these results align the observed efficiency gains with the two-stage design of AtlasNav: persistent multi-view organization supplies reusable directions before interaction, and query-adaptive navigation determines which of them to exploit first. Their combination reduces Evidence Blindness earlier, accelerates convergence toward the evidence-supplied reference, and ultimately improves the accuracy--efficiency frontier.

\subsection{Robustness to Corpus Structure and Scale Shifts}
\label{sec:phantomwiki_results}
PhantomWiki \citep{gong2025phantomwiki} tests whether AtlasNav transfers beyond the corpus geometry of web-style deep research. Its shorter, entity-centered documents are connected by relational and attribute chains, producing a substantially different navigation structure from BrowseComp-Plus. This setting therefore tests whether persistent multi-view navigation remains effective under both structural and scale shifts.

Figure~\ref{fig:phantomwiki_surface} shows that AtlasNav maintains the lowest Surface Evidence Blindness at every PhantomWiki scale, including the million-file setting. The persistent multi-view Atlas therefore continues to expose useful evidence directions even when corpus organization shifts from web-style documents to entity-centered relational chains. This suggests that its evidence-access advantage does not depend on a single corpus geometry.

Table~\ref{tab:phantomwiki_accuracy} shows that this process-level advantage carries over to task accuracy: AtlasNav achieves the highest strict accuracy at every scale. Notably, DCI consistently outperforms DR-DCI on PhantomWiki, reversing their ordering on BrowseComp-Plus. This reversal suggests that query-conditioned workspace reconstruction is sensitive to corpus organization, whereas AtlasNav is less dependent on the geometry encountered at query time because its navigation structure is built persistently in advance.

Across all scales, Surface Evidence Blindness and strict accuracy show the same ordering among interfaces, linking better evidence exposure to better task performance under this structural shift. Scaling from 100K to roughly one million files degrades all methods as the distractor space grows, but AtlasNav retains both the lowest Surface blindness and the highest accuracy. Persistent organization therefore remains useful as the navigation space expands, while not eliminating the intrinsic difficulty of very large corpora.

\begingroup
\setlength{\intextsep}{7pt plus 1pt minus 1pt}
\setlength{\textfloatsep}{0pt plus 0pt minus 0pt}
\setlength{\abovecaptionskip}{1pt}
\setlength{\belowcaptionskip}{0pt}
\begin{figure}[!t]
  \centering
  \begin{minipage}[t]{0.47\textwidth}
    \vspace{0pt}
    \centering
    \includegraphics[width=0.92\linewidth]{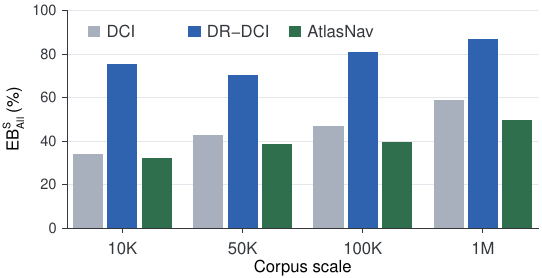}
    \vspace{-4pt}
    \normalsize
    \caption{$\operatorname{EB}_{\mathrm{All}}^{S}$ under controlled PhantomWiki scaling. Questions and supporting evidence remain fixed; lower is better.}
    \label{fig:phantomwiki_surface}
  \end{minipage}
  \hfill
  \begin{minipage}[t]{0.50\textwidth}
    \vspace{0pt}
    \makeatletter
    \def\@captype{table}
    \makeatother
    \caption{PhantomWiki accuracy under controlled corpus scaling ($n=200$). Bold denotes the best result.}
    \label{tab:phantomwiki_accuracy}
    \centering
    \small
    \setlength{\tabcolsep}{2.8pt}
    \renewcommand{\arraystretch}{1.37}
    \begin{tabular*}{\linewidth}{@{\extracolsep{\fill}}l|rrrr@{}}
      \toprule
      & \multicolumn{4}{c}{Strict Accuracy (\%) $\uparrow$} \\
      \cmidrule(l){2-5}
      Interface & 10K & 50K & 100K & 1M \\
      \midrule
      DCI      & 70.0 & 72.5 & 70.0 & 58.5 \\
      DR-DCI   & 41.5 & 45.5 & 45.0 & 37.0 \\
      \textbf{AtlasNav} & \textbf{76.5} & \textbf{77.5} & \textbf{76.0} & \textbf{61.0} \\
      \bottomrule
    \end{tabular*}
  \end{minipage}
  \vspace{-10pt}
\end{figure}

\FloatBarrier
\endgroup
\subsection{Transfer to Heterogeneous Enterprise Knowledge}
\label{sec:enterprise_results}
EnterpriseRAG-Bench tests whether AtlasNav transfers from web-style and synthetic corpora to heterogeneous enterprise knowledge. Without source-specific redesign, AtlasNav applies the same persistent Topic, Identity, Episode, and Relation organization across diverse enterprise sources.

Table~\ref{tab:enterprise_main} shows strong transfer: AtlasNav achieves the highest Overall score of 87.00, exceeding the strongest listed public system by 0.42 points.

\begingroup
\setlength{\intextsep}{2pt plus 0.5pt minus 0.5pt}
\setlength{\abovecaptionskip}{0pt}
\setlength{\belowcaptionskip}{2pt}
\begin{table}[H]
  \caption{EnterpriseRAG-Bench metrics with representative public systems.}
  \label{tab:enterprise_main}
  \centering
  \small
  \setlength{\tabcolsep}{2.5pt}
  \renewcommand{\arraystretch}{1.00}
  \begin{tabular*}{\textwidth}{@{\extracolsep{\fill}}l|ccccc@{}}
    \toprule
    System & Overall $\uparrow$ & Correct. (\%) $\uparrow$ & Complete. (\%) $\uparrow$
      & Doc. Recall (\%) $\uparrow$ & Invalid Extra $\downarrow$ \\
    \midrule
    \textbf{AtlasNav}             & \textbf{87.00} & \textbf{94.40} & 89.14 & 87.05 & 0.66 \\
    Mixedbread (+ Opus 5)        & 86.58 & 89.80 & \textbf{90.62} & \textbf{94.28} & \textbf{0.39} \\
    Prism (aurait.ai)            & 80.46 & 85.60 & 86.26 & 87.81 & 3.31 \\
    metor.com                    & 80.34 & 82.00 & 86.22 & 85.53 & 4.96 \\
    ZNV AgentCube                & 80.26 & 82.40 & 86.56 & 86.10 & 2.94 \\
    Skyller AI                   & 79.30 & 81.60 & 87.39 & 86.50 & 8.74 \\
    CDL (Causal Dynamics Lab)    & 78.95 & 82.00 & 85.16 & 80.54 & 14.08 \\
    \bottomrule
  \end{tabular*}
\end{table}
\endgroup

AtlasNav combines the highest Correctness (94.40\%) with strong Completeness (89.14\%). It also achieves 87.05\% Document Recall with only 0.66 Invalid Extra Documents, showing that its selected evidence covers most required documents while introducing few unnecessary ones. Together, these results indicate that AtlasNav can maintain both answer quality and focused evidence selection across heterogeneous enterprise sources.

Category-level results reveal a clear boundary. AtlasNav performs strongly on intra-document reasoning (96.46\% Overall), constrained queries (94.57\%), and conflicting-information cases (92.69\%), where evidence is relatively localized. It is weaker on project-related (59.09\%) and completeness-oriented (66.98\%) questions that require broader aggregation of dispersed evidence, indicating that exhaustive cross-document coverage remains the main challenge. Even so, achieving the highest Overall score across heterogeneous enterprise data while remaining strong across several reasoning patterns shows that the same persistent navigation principle generalizes well beyond the web-style and synthetic settings used elsewhere in the paper.

%% file: sections/05_conclusion.tex
\section{Conclusion}
\label{sec:conclusion}

\emph{Interactive corpus access} is limited not only by whether evidence is available, but by whether agents can make that evidence usable within a finite interaction budget. We formalize this problem as Evidence Blindness and introduce AtlasNav, which organizes reusable multi-view corpus structure before query time and supports adaptive navigation during interaction. On BrowseComp-Plus, AtlasNav reduces evidence loss, improves accuracy, and lowers inference cost; its advantages also persist under large corpus-scale shifts and heterogeneous enterprise data. More broadly, our results show that effective interactive search depends not only on access to evidence, but also on how the corpus is organized to guide limited interaction toward it.

%% file: sections/06_statements.tex
\clearpage
\subsection*{AI use statement}

In this work, we used generative AI tools for language editing, notation and
reference-format checks, software-code assistance, \LaTeX{} and figure/table
layout checks, and targeted diagnostic review of selected model-visible
trajectories and reported tables. We did not use generative AI to determine the
research questions, formulate the method, design the experimental protocol,
alter benchmark annotations or evaluation references, generate or reinterpret
reported outcomes, or make final analytical or conclusion decisions. The
authors reviewed all AI-assisted text for fidelity to the intended claims and
supporting evidence, inspected and tested AI-assisted code against fixed inputs
and expected outputs, and independently checked all diagnostic suggestions.
The authors take responsibility for all text, claims, code, analyses, and
artifacts in this submission.

\subsection*{Reproducibility statement}

Sections~\ref{sec:evidence_blindness}--\ref{sec:experiments} describe the
Evidence Blindness evaluation, AtlasNav, and the main experiments.
Appendix~\ref{app:atlasnav_implementation} specifies the executable evidence
stages and frozen fragment-level reference and documents Atlas construction,
routing, and online navigation; Appendix~\ref{app:experimental_details} gives
the evaluation, cost, checkpoint, and statistical protocols. Exact endpoint and
checkpoint values are reported in Appendix~\ref{app:complete_main_results}.
The anonymized supplementary package contains the provider-neutral AtlasNav
runtime, the DCI-harness compatibility source and patch metadata, configurations,
tests, documentation, prompts, licences, compact result records and diagnostics,
the BrowseComp evidence-span qrel, EnterpriseRAG evaluation records, and
checksummed provenance manifests. \texttt{ENVIRONMENT.md} specifies the Python 3.11+
installation and offline validation procedure (27 tests), as well as the
optional historical Node.js/pi-mono path and its recorded external commit.
Because the compact package omits the large Atlas maps, query embeddings, full
trajectory archives, and upstream corpora, it supports code/configuration
inspection, compact result checks, and reruns with independently obtained inputs
but does not by itself provide exact standalone replay of every historical
trajectory. No credentials or machine-local paths are included.




%% file: sections/appendix.tex
\clearpage
\section{Limitations}
\label{app:limitations}

\noindent\textbf{Extreme-scale navigation remains challenging.}
Although AtlasNav retains an advantage as PhantomWiki grows, its own
performance still declines at the largest scale. Persistent organization
therefore mitigates rather than removes the difficulty of increasingly large
distractor spaces. More aggressive multi-resolution navigation, adaptive region
expansion, or scalable pruning may be needed at substantially larger scales.

\noindent\textbf{The current Atlas is designed primarily for static or slowly
evolving corpora.}
Efficiently maintaining neighborhoods, hierarchy, anchors, and cross-region
links under continuous corpus changes is not studied here. In evolving
deployments, the Atlas may therefore need periodic reconstruction or incremental
maintenance, with its persistent structure reused between updates.

\noindent\textbf{Persistent organization also does not eliminate downstream
evidence-access failures.}
Evidence may still remain unopened, unlocalized, or incorrectly verified and
synthesized after being surfaced. Evidence Blindness should therefore be
interpreted as a diagnostic of evidence realization under a fixed interaction
setting, not as a complete characterization of all failure sources. Its Locate
checkpoint uses a fragment-level reference fixed before evaluation, which
provides a consistent target but does not enumerate every semantically
equivalent evidence path.

\noindent\textbf{Finally, our efficiency analysis focuses on query-time agent
inference cost.}
AtlasNav additionally requires offline corpus preparation, including
representation construction and indexing. Because these costs depend on
implementation and deployment conditions, the reported online-cost frontier is
not a complete lifecycle-cost comparison; the practical trade-off depends on
corpus reuse, update frequency, and deployment setting.

\input{sections/appendix_b_complete}

\input{sections/appendix_c_protocols}

\input{sections/appendix_d_method_details}

\input{sections/appendix_e_main_analysis}

\input{sections/appendix_f_boundary}

\input{sections/appendix_g_prompts}

%% file: sections/appendix_b_complete.tex
\section{Complete Results for Main Experiments}
\label{app:complete_main_results}

This appendix reports the supplementary numerical results underlying the three
main experiments in Section~\ref{sec:experiments}. We provide the endpoint values
and recorded budget-checkpoint results underlying the summaries and figures in
the main text; interpretation and additional analyses are deferred to
Appendix~\ref{app:complete_results}.

\subsection{BrowseComp-Plus}

Tables~\ref{tab:app_main_endpoint_eb}--\ref{tab:app_main_cost_checkpoints}
report supplementary BrowseComp-Plus endpoint and budget-checkpoint results.
Bold marks the best interface within each comparison group.

\begin{table}[!htbp]
\caption{Mean endpoint Evidence Blindness on BrowseComp-Plus (\%); lower is
better. Any-slot and All-slot results are reported in the main text.
Construction is omitted because it is saturated for all evaluated interfaces.}
\label{tab:app_main_endpoint_eb}
\centering
\footnotesize
\setlength{\tabcolsep}{6pt}
\renewcommand{\arraystretch}{0.95}
\begin{tabular*}{\textwidth}{@{\extracolsep{\fill}}llrrr@{}}
\toprule
Backbone & Interface & Surface & Open & Locate \\
\midrule
GPT-5.6-Luna & DCI & 6.81 & 12.93 & 16.02 \\
& DR-DCI & 9.58 & 10.78 & 16.27 \\
& AtlasNav & \textbf{5.84} & \textbf{9.28} & \textbf{12.29} \\
\midrule
DeepSeek & DCI & 13.25 & 15.04 & 18.29 \\
& DR-DCI & 13.80 & 16.81 & 24.04 \\
& AtlasNav & \textbf{4.90} & \textbf{7.73} & \textbf{11.08} \\
\midrule
Qwen & DCI & 55.12 & 58.07 & 62.53 \\
& DR-DCI & 49.40 & 50.96 & 56.93 \\
& AtlasNav & \textbf{19.90} & \textbf{22.65} & \textbf{29.86} \\
\midrule
MiMo & DCI & 29.52 & 33.86 & 36.87 \\
& DR-DCI & 26.10 & 27.31 & 32.01 \\
& AtlasNav & \textbf{17.35} & \textbf{20.00} & \textbf{23.98} \\
\bottomrule
\end{tabular*}
\end{table}

\begin{table}[!htbp]
\caption{BrowseComp-Plus results at turn checkpoints. Panel (a) reports strict
accuracy (\%); panel (b) reports checkpoint-wise
$\operatorname{EB}_{\mathrm{All}}^{S/O/L}$ (\%), with each entry as Surface /
Open / Locate.}
\label{tab:app_main_turn_checkpoints}
\centering
\footnotesize
\setlength{\tabcolsep}{5pt}
\renewcommand{\arraystretch}{0.95}
\textbf{(a) Strict Accuracy ($\uparrow$)}\\[1pt]
\begin{tabular*}{\textwidth}{@{\extracolsep{\fill}}llrrrrr@{}}
\toprule
Backbone & Interface & 15 & 30 & 60 & 120 & 300 \\
\midrule
GPT-5.6-Luna & DCI & 36.75 & 69.64 & 83.73 & 87.35 & 87.35 \\
& DR-DCI & 46.14 & 69.76 & 80.96 & 85.78 & 87.83 \\
& AtlasNav & \textbf{50.00} & \textbf{76.51} & \textbf{89.04} & \textbf{91.81} & \textbf{91.81} \\
\midrule
DeepSeek & DCI & 21.20 & 41.33 & 58.19 & 72.29 & 81.45 \\
& DR-DCI & 30.00 & 50.48 & 68.55 & 78.80 & 84.34 \\
& AtlasNav & \textbf{41.57} & \textbf{65.06} & \textbf{80.84} & \textbf{90.36} & \textbf{92.05} \\
\midrule
Qwen & DCI & 8.43 & 14.58 & 25.18 & 33.98 & 36.99 \\
& DR-DCI & \textbf{15.30} & 28.92 & 40.12 & 47.71 & 50.96 \\
& AtlasNav & 14.58 & \textbf{37.71} & \textbf{56.02} & \textbf{66.99} & \textbf{72.53} \\
\midrule
MiMo & DCI & 16.51 & 29.40 & 43.98 & 55.18 & 62.89 \\
& DR-DCI & 17.59 & 37.95 & 55.42 & 67.23 & 71.08 \\
& AtlasNav & \textbf{22.17} & \textbf{45.06} & \textbf{61.93} & \textbf{72.05} & \textbf{79.04} \\
\bottomrule
\end{tabular*}

\vspace{4pt}
\textbf{(b) $\operatorname{EB}_{\mathrm{All}}^{S/O/L}$ ($\downarrow$), turns 15--300}\\[1pt]
\scriptsize
\setlength{\tabcolsep}{1.5pt}
\begin{tabular*}{\textwidth}{@{\extracolsep{\fill}}llrrrrr@{}}
\toprule
Backbone & Interface & 15 & 30 & 60 & 120 & 300 \\
\midrule
GPT-5.6-Luna & DCI & 27.23/39.40/44.46 & 12.65/21.45/25.30 & 8.07/14.70/18.31 & 6.99/13.01/16.27 & 6.99/13.01/16.27 \\
& DR-DCI & \textbf{23.86}/\textbf{27.59}/38.92 & 16.02/18.43/27.11 & 12.41/13.98/20.72 & 10.24/11.69/17.71 & 9.76/11.08/16.63 \\
& AtlasNav & 25.30/33.01/\textbf{36.99} & \textbf{12.17}/\textbf{17.59}/\textbf{20.48} & \textbf{6.75}/\textbf{10.48}/\textbf{13.61} & \textbf{6.02}/\textbf{9.52}/\textbf{12.65} & \textbf{6.02}/\textbf{9.52}/\textbf{12.65} \\
\midrule
DeepSeek & DCI & 50.84/59.04/62.77 & 38.07/43.37/48.07 & 24.94/28.67/33.49 & 17.95/20.72/24.82 & 13.37/15.42/18.80 \\
& DR-DCI & 35.54/43.13/50.48 & 25.06/31.08/39.64 & 18.19/22.65/30.84 & 15.42/18.92/26.63 & 14.22/17.23/24.58 \\
& AtlasNav & \textbf{25.42}/\textbf{32.77}/\textbf{38.19} & \textbf{14.10}/\textbf{18.43}/\textbf{22.77} & \textbf{7.35}/\textbf{10.84}/\textbf{14.58} & \textbf{4.94}/\textbf{8.07}/\textbf{11.81} & \textbf{4.94}/\textbf{7.83}/\textbf{11.45} \\
\midrule
Qwen & DCI & 79.40/81.69/86.02 & 71.57/73.13/78.07 & 63.61/64.34/70.36 & 58.67/60.48/65.54 & 55.54/58.80/63.49 \\
& DR-DCI & 64.34/67.23/73.73 & 59.52/62.41/67.47 & 54.58/56.63/62.29 & 51.57/53.25/59.40 & 49.76/51.33/57.23 \\
& AtlasNav & \textbf{56.87}/\textbf{62.53}/\textbf{69.76} & \textbf{41.20}/\textbf{46.27}/\textbf{53.86} & \textbf{28.19}/\textbf{32.53}/\textbf{40.96} & \textbf{22.41}/\textbf{25.42}/\textbf{33.13} & \textbf{20.12}/\textbf{23.01}/\textbf{30.36} \\
\midrule
MiMo & DCI & 57.83/67.47/71.20 & 48.55/56.51/59.64 & 39.16/45.42/48.92 & 33.13/38.07/41.45 & 29.52/33.86/36.87 \\
& DR-DCI & 53.98/55.78/62.41 & 42.41/43.98/50.48 & 33.01/34.58/40.00 & 27.95/29.40/34.34 & 26.27/27.59/32.29 \\
& AtlasNav & \textbf{46.27}/\textbf{52.41}/\textbf{57.95} & \textbf{34.10}/\textbf{39.16}/\textbf{44.34} & \textbf{25.54}/\textbf{29.64}/\textbf{33.86} & \textbf{20.60}/\textbf{24.34}/\textbf{28.19} & \textbf{17.83}/\textbf{20.24}/\textbf{24.22} \\
\bottomrule
\end{tabular*}

\vspace{3pt}
{\footnotesize\raggedright
Endpoint values need not coincide with the 300-turn checkpoint because endpoint
evaluation follows the frozen finalization protocol, whereas checkpoints use
recorded trajectory prefixes.\par}
\end{table}

\makeatletter
\setlength{\@fptop}{0pt}
\setlength{\@fpsep}{12pt}
\setlength{\@fpbot}{0pt plus 1fil}
\makeatother

\begin{table}[!htbp]
\caption{BrowseComp-Plus results at within-backbone query-time agent-cost
checkpoints. Each row is one cost budget for one backbone; the first block
reports strict accuracy (\%) and the second checkpoint-wise
$\operatorname{EB}_{\mathrm{All}}^{S/O/L}$ (\%), with each EB entry as Surface /
Open / Locate. Lower is better for EB.}
\label{tab:app_main_cost_checkpoints}
\centering
\scriptsize
\setlength{\tabcolsep}{2.5pt}
\renewcommand{\arraystretch}{0.95}
\begin{tabular*}{\textwidth}{@{\extracolsep{\fill}}ll|ccc|ccc@{}}
\toprule
& & \multicolumn{3}{c|}{Strict Accuracy (\%) $\uparrow$}
  & \multicolumn{3}{c}{$\operatorname{EB}_{\mathrm{All}}^{S/O/L}$ (\%) $\downarrow$} \\
\cmidrule(lr){3-5}\cmidrule(l){6-8}
Backbone & Cost & DCI & DR-DCI & AtlasNav & DCI & DR-DCI & AtlasNav \\
\midrule
GPT-5.6-Luna & 0.025 & 45.66 & 65.42 & \textbf{66.39} & 24.70/36.14/40.60 & 17.47/\textbf{20.12}/29.88 & \textbf{16.51}/23.61/\textbf{26.75} \\
& 0.05 & 57.83 & 74.82 & \textbf{75.90} & 20.24/30.00/34.46 & 14.10/\textbf{16.02}/23.73 & \textbf{13.37}/18.67/\textbf{21.57} \\
& 0.15 & 69.52 & 81.20 & \textbf{82.05} & 13.49/22.77/26.51 & 12.77/\textbf{14.46}/20.84 & \textbf{10.12}/15.30/\textbf{18.07} \\
& 0.65 & 84.46 & 86.39 & \textbf{91.81} & 7.95/14.58/18.07 & 10.12/11.57/17.47 & \textbf{6.02}/\textbf{9.52}/\textbf{12.65} \\
\midrule
DeepSeek & 0.15 & 49.16 & 50.36 & \textbf{70.96} & 43.73/49.88/54.46 & 29.40/36.02/45.18 & \textbf{23.25}/\textbf{30.36}/\textbf{35.90} \\
& 0.40 & 62.17 & 65.54 & \textbf{82.89} & 31.69/37.35/42.05 & 20.48/25.66/34.46 & \textbf{12.29}/\textbf{16.39}/\textbf{20.96} \\
& 1.20 & 70.96 & 79.76 & \textbf{89.16} & 21.57/24.82/29.52 & 16.63/20.48/28.43 & \textbf{6.87}/\textbf{10.24}/\textbf{13.73} \\
& 2.75 & 76.99 & 82.77 & \textbf{92.05} & 16.87/19.52/23.49 & 15.30/18.55/26.14 & \textbf{4.94}/\textbf{7.83}/\textbf{11.45} \\
\midrule
Qwen & 0.15 & 12.17 & 29.16 & \textbf{40.36} & 77.35/79.64/83.98 & 60.00/62.65/67.47 & \textbf{37.35}/\textbf{42.65}/\textbf{51.20} \\
& 0.40 & 16.51 & 39.40 & \textbf{53.01} & 70.60/72.17/76.99 & 54.58/56.75/62.89 & \textbf{29.16}/\textbf{33.61}/\textbf{42.41} \\
& 0.80 & 19.76 & 46.14 & \textbf{63.86} & 67.47/69.76/71.81 & 51.93/53.73/59.76 & \textbf{24.46}/\textbf{28.19}/\textbf{36.02} \\
& 1.20 & 24.70 & 48.80 & \textbf{67.47} & 63.25/64.82/70.84 & 50.48/52.17/58.55 & \textbf{22.17}/\textbf{25.42}/\textbf{33.13} \\
& 1.95 & 28.55 & 50.72 & \textbf{72.53} & 60.48/62.77/67.47 & 50.00/51.57/57.47 & \textbf{20.12}/\textbf{23.01}/\textbf{30.36} \\
\midrule
MiMo & 0.15 & 47.35 & 62.05 & \textbf{65.54} & 45.66/52.77/55.90 & 33.37/35.30/40.96 & \textbf{26.75}/\textbf{30.84}/\textbf{35.54} \\
& 0.40 & 51.81 & 65.78 & \textbf{69.04} & 40.84/47.23/50.48 & 30.00/31.45/36.63 & \textbf{24.82}/\textbf{28.92}/\textbf{33.01} \\
& 0.80 & 54.82 & 67.23 & \textbf{71.08} & 37.95/43.37/46.99 & 28.19/29.64/34.70 & \textbf{22.41}/\textbf{26.27}/\textbf{30.48} \\
& 1.20 & 56.99 & 68.80 & \textbf{76.75} & 34.70/40.12/43.61 & 27.47/28.92/34.10 & \textbf{19.88}/\textbf{23.73}/\textbf{27.95} \\
& 1.40 & 57.71 & 70.72 & \textbf{79.04} & 30.96/35.54/39.04 & 26.27/27.59/32.29 & \textbf{17.83}/\textbf{20.24}/\textbf{24.22} \\
\bottomrule
\end{tabular*}

\vspace{3pt}
{\scriptsize\raggedright
Costs are in USD for GPT-5.6-Luna and in CNY for all other backbones. DeepSeek
and MiMo use active checkpoint finalization; intermediate GPT-5.6-Luna and Qwen
values are natural-terminal lower bounds.\par}
\end{table}

\FloatBarrier
\makeatletter
\setlength{\@fptop}{0pt plus 1fil}
\setlength{\@fpsep}{8pt plus 2fil}
\setlength{\@fpbot}{0pt plus 1fil}
\makeatother

\setlength{\intextsep}{8pt plus 2pt minus 2pt}
\setlength{\textfloatsep}{8pt plus 2pt minus 2pt}

\subsection{PhantomWiki}

Table~\ref{tab:app_main_phantom} reports the downstream evidence-realization
results omitted from the main text.

\begin{table}[!htbp]
\caption{PhantomWiki evidence realization beyond the main text. Questions and
supporting evidence are fixed across nested corpus sizes. Surface-to-Open is
the fraction of complete-Surface questions that also reach complete Open.}
\label{tab:app_main_phantom}
\centering
\small
\setlength{\tabcolsep}{6pt}
\renewcommand{\arraystretch}{1.00}
\begin{tabular*}{\textwidth}{@{\extracolsep{\fill}}rrlrr@{}}
\toprule
Scale & Files & Interface & $\operatorname{EB}_{\mathrm{All}}^{O}$ $\downarrow$
  & Surface-to-Open (\%) $\uparrow$ \\
\midrule
10K & 10,059 & DCI & 38.0 & 93.94 \\
& & DR-DCI & 76.5 & 95.92 \\
& & AtlasNav & 45.5 & 80.15 \\
\midrule
50K & 50,324 & DCI & 52.0 & 84.21 \\
& & DR-DCI & 76.5 & 79.66 \\
& & AtlasNav & 52.5 & 77.24 \\
\midrule
100K & 100,647 & DCI & 58.5 & 78.30 \\
& & DR-DCI & 85.5 & 76.32 \\
& & AtlasNav & 50.5 & 81.82 \\
\midrule
1M & 1,006,839 & DCI & 68.5 & 76.83 \\
& & DR-DCI & 89.5 & 80.77 \\
& & AtlasNav & 71.0 & 57.43 \\
\bottomrule
\end{tabular*}
\end{table}

\FloatBarrier
\subsection{EnterpriseRAG-Bench}

Tables~\ref{tab:app_main_enterprise_types} and
\ref{tab:app_main_enterprise_sources} report AtlasNav's complete category- and
source-level metrics; the overall comparison with representative public systems
is given in Table~\ref{tab:enterprise_main}.

\begin{table}[!htbp]
\caption{Complete AtlasNav metrics by EnterpriseRAG-Bench question category.}
\label{tab:app_main_enterprise_types}
\centering
\small
\setlength{\tabcolsep}{4pt}
\renewcommand{\arraystretch}{1.00}
\begin{tabular*}{\textwidth}{@{\extracolsep{\fill}}lrrrrrr@{}}
\toprule
Category & $n$ & Overall $\uparrow$ & Correct. $\uparrow$ & Complete. $\uparrow$
  & Doc. Recall $\uparrow$ & Invalid Extra $\downarrow$ \\
\midrule
Basic & 175 & 95.16 & 98.86 & 95.16 & 93.71 & 0.26 \\
Semantic & 125 & 80.73 & 89.60 & 82.25 & 80.00 & 0.80 \\
Intra-document reasoning & 40 & 96.46 & 100.00 & 96.46 & 97.50 & 0.05 \\
Project related & 40 & 59.09 & 87.50 & 67.88 & 65.48 & 2.77 \\
Constrained & 30 & 94.57 & 96.67 & 96.79 & 95.00 & 0.70 \\
Conflicting information & 20 & 92.69 & 100.00 & 92.69 & 80.00 & 0.20 \\
Completeness & 20 & 66.98 & 80.00 & 75.06 & 77.15 & 1.30 \\
Miscellaneous & 20 & 96.08 & 100.00 & 96.08 & 100.00 & 0.05 \\
High level & 10 & 58.00 & 70.00 & 88.00 & -- & -- \\
Information not found & 20 & 100.00 & 100.00 & 100.00 & -- & -- \\
\bottomrule
\end{tabular*}

\vspace{3pt}
{\footnotesize\raggedright
Document Recall and Invalid Extra are defined for the 470 questions with
expected documents, matching the official evaluator.\par}
\end{table}

\begin{table}[!htbp]
\caption{Complete AtlasNav metrics by overlapping EnterpriseRAG-Bench source
tag. Counts do not sum to the benchmark size because questions may have
multiple source tags.}
\label{tab:app_main_enterprise_sources}
\centering
\small
\setlength{\tabcolsep}{4pt}
\renewcommand{\arraystretch}{1.00}
\begin{tabular*}{\textwidth}{@{\extracolsep{\fill}}lrrrrrr@{}}
\toprule
Source tag & $n$ & Overall $\uparrow$ & Correct. $\uparrow$ & Complete. $\uparrow$
  & Doc. Recall $\uparrow$ & Invalid Extra $\downarrow$ \\
\midrule
Confluence & 115 & 78.00 & 91.30 & 81.68 & 80.92 & 1.37 \\
Fireflies & 24 & 87.22 & 95.83 & 87.22 & 74.58 & 0.38 \\
GitHub & 60 & 82.80 & 93.33 & 86.07 & 82.67 & 0.98 \\
Gmail & 55 & 86.74 & 94.55 & 89.58 & 86.00 & 0.45 \\
Google Drive & 59 & 88.70 & 96.61 & 89.53 & 83.92 & 1.05 \\
HubSpot & 33 & 92.42 & 96.97 & 92.93 & 96.97 & 0.03 \\
Jira & 100 & 81.21 & 96.00 & 83.43 & 81.52 & 1.03 \\
Linear & 56 & 84.23 & 96.43 & 86.47 & 88.81 & 0.96 \\
Slack & 78 & 80.03 & 87.18 & 84.39 & 74.22 & 1.09 \\
\bottomrule
\end{tabular*}

\vspace{3pt}
{\footnotesize\raggedright
Document Recall and Invalid Extra are defined for the questions with expected
documents.\par}
\end{table}

\FloatBarrier

\setlength{\intextsep}{12pt plus 2pt minus 2pt}
\setlength{\textfloatsep}{6pt plus 1pt minus 1pt}

%% file: sections/appendix_c_protocols.tex
\section{Experimental Setup and Evaluation Protocols}
\label{app:experimental_details}

This appendix provides experimental and evaluation details omitted from
Section~\ref{sec:experimental_setup}. Supplementary numerical results are
reported in Appendix~\ref{app:complete_main_results}, while implementation
details for Evidence Blindness and AtlasNav are provided in
Appendix~\ref{app:atlasnav_implementation}.

\subsection{Experimental Setup}
\label{app:controlled_setup}

Across controlled comparisons, interfaces use the same language-model
backbone, benchmark denominator, and nominal interaction ceiling, while
retaining access to the original corpus. AtlasNav does not obtain its
advantage by permanently pruning documents; original documents remain
directly accessible throughout interaction.

\paragraph{Implementation control.}
AtlasNav is implemented on top of the open-sourced DCI interaction harness,
which we retain as the common agent execution framework for the controlled
comparisons. We keep the shared interaction loop, tool invocation protocol, and
evaluation pipeline fixed, and modify the corpus-access layer to expose
persistent Atlas-guided navigation. This isolates the effect of corpus
organization from differences in the underlying agent harness.

For BrowseComp-Plus, we additionally construct a same-model
evidence-supplied empirical reference. The model receives benchmark-designated
gold documents but not the reference answer, largely bypassing evidence
discovery. This reference is used only to characterize evidence-conditioned
capability and is not treated as a theoretical upper bound or deployable
baseline.

EnterpriseRAG-Bench follows its official Overall, Correctness, Completeness,
Document Recall, and Invalid Extra metrics. Correctness measures final-answer
correctness, Completeness measures required-information coverage, Document
Recall measures coverage of required supporting documents, and Invalid Extra
counts unnecessary selected documents; Overall follows the benchmark's
official aggregate score.

EnterpriseRAG-Bench leaderboard results reported in
Table~\ref{tab:enterprise_main} are based on the public leaderboard snapshot
collected on September 24, 2026, at 23:59 (UTC time). We use this fixed
snapshot to ensure reproducibility, as the online leaderboard may continue to
change after submission.

\subsection{Trajectory and Budget Protocol}
\label{app:trajectory_protocol}

Each question contributes exactly one evaluation trajectory. We use no
best-of-$N$ selection, correctness-conditioned trajectory replacement, or
error-targeted reruns; invalid or failed runs remain in the benchmark
denominator and are scored as incorrect.

Turn and cost checkpoints are computed from frozen trajectory prefixes.
Evidence or answers appearing after a checkpoint are never backfilled, so each
checkpoint reflects only the state naturally realized within that budget.
Turn checkpoints measure interaction depth, while cost checkpoints use
accumulated query-time agent inference cost; because provider pricing and
currencies differ, cost comparisons are made only within backbone.

To control long-tail search expansion, we use frozen backbone-specific
finalization triggers: CNY~2.75 for DeepSeek, CNY~1.40 for MiMo, USD~0.65 for
GPT-5.6-Luna, and CNY~1.95 for Qwen.
\label{app:finalization_protocol}
Once a trigger is reached, the harness asks the agent to stop broadening
search and finalize from available evidence. Because provider usage is
observed only after a response completes, limited overshoot may occur.
Finalization depends only on usage, turn count, tool state, and final-answer
validity; it never reads gold evidence, Qrels, judge outcomes, or correctness.
All interfaces using the same backbone follow the same frozen rule.

\subsection{Cost Accounting}
\label{app:cost_accounting}

Query-time agent inference cost includes model inference used for searching,
reading, reasoning, and producing the final answer. Post-hoc judging, offline
corpus preparation, and query embedding are excluded.

For an Atlas shared across $N$ queries, we separate reusable preparation from
per-query operating cost:
\begin{equation}
\bar C_A(N)
=
C_A^{\mathrm{op}}
+
\frac{C_A^{\mathrm{reuse}}}{N},
\end{equation}
where $C_A^{\mathrm{op}}$ is the per-query operating cost and
$C_A^{\mathrm{reuse}}$ is reusable corpus-preparation cost. Detailed cost
distributions and amortization analyses are reported in
Appendix~\ref{app:offline_resources}.

\subsection{Statistics and Reproducibility}
\label{app:statistics_reproducibility}

For paired binary outcomes, we use exact McNemar tests; paired mean
differences use fixed-seed bootstrap confidence intervals. We do not infer
statistical significance from aggregate-only results when per-instance
records are unavailable.

Corpus representations, Atlas memberships, router parameters, evaluation
assets, and reported trajectories are frozen before analysis. Versioned
records associate each result with its Qrel, model configuration, trajectory
termination, judge outcome, and checkpoint data.

%% file: sections/appendix_d_method_details.tex
\section{Method Details}
\label{app:atlasnav_implementation}
\label{app:evidence_blindness}

This appendix complements Sections~\ref{sec:evidence_blindness}
and~\ref{sec:method} with the operational details required to reproduce
Evidence Blindness and AtlasNav. We first specify how the four
evidence-realization stages are instantiated from model-visible trajectories
and how Locate is measured with a frozen fragment-level reference. We then
describe the persistent multi-view Atlas, query-adaptive routing, and the
online navigation interface.

\subsection{Operationalizing Evidence Blindness}
\label{app:evidence_predicates}

Section~\ref{sec:evidence_blindness} defines Evidence Blindness over the
required evidence slots $\mathcal{E}_q$ and their realized subsets
$\mathcal{E}_q^\tau$ at checkpoints $\tau\in\{C,S,O,L\}$. Here we specify the
executable trajectory conditions used to determine whether a slot is realized
at each checkpoint.

\paragraph{Construction.}
A slot is realized at Construction if valid supporting evidence for that slot
exists in the original corpus accessible through the interface. In the main
BrowseComp-Plus comparisons, Construction is saturated because all annotated
supporting evidence is present and no evaluated interface permanently removes
the corresponding documents from access.

\paragraph{Surface.}
A slot is realized at Surface when an identifiable entry to one of its
supporting documents appears in an observation actually shown to the model.
Internal retrieval state or metadata that never enters the model-visible
observation does not count.

\paragraph{Open.}
A slot is realized at Open when the original content of a supporting document
enters model-visible context through a successful document read. Filenames,
paths, region labels, previews, or failed tool outputs are insufficient.

\paragraph{Locate.}
A slot is realized at Locate when an accepted decision-relevant fragment
aligned to that slot enters model-visible context. The fragment-level reference
and matching procedure are described in
Appendix~\ref{app:locate_construct}.

Trajectory events are processed in temporal order. At any turn or cost
checkpoint, only evidence realized within the corresponding frozen trajectory
prefix is counted; later observations are never backfilled. These rules
preserve the monotonic realization funnel defined in
Section~\ref{sec:evidence_blindness}.

\subsection{Fragment-Level Reference for Locate}
\label{app:locate_construct}
\label{app:qrel_construction}
\label{app:evidence_diagnostics}

Surface and Open can be determined from document-level trajectory events,
whereas Locate requires a content-level reference. We therefore construct a
fragment-level reference over the full BrowseComp-Plus evaluation set
\citep{chen2025browsecompplus} before evaluation. Its purpose is to provide a
deterministic target for deciding whether the
decisive evidence associated with a required slot has entered model-visible
context.

The reference is built from benchmark questions, reference answers, document
supervision, and original document text \citep{chen2025browsecompplus},
independently of evaluated trajectories, model predictions, judge outcomes,
and correctness labels. For each required evidence slot, we retain one or more
short fragments that directly establish the corresponding factual constraint;
when multiple supporting documents or passages are valid, all aligned
alternatives are preserved.

\begin{table}[!htbp]
\caption{Frozen BrowseComp-Plus fragment-level Qrel used to operationalize
Answer-Evidence Locate.}
\label{tab:app_qrel_stats}
\centering
\small
\begin{tabular*}{.72\textwidth}{@{\extracolsep{\fill}}lr@{}}
\toprule
Item & Count \\
\midrule
Questions & 830 \\
Required evidence slots & 889 \\
Accepted canonical fragments & 1,753 \\
Canonical documents represented & 1,445 \\
Questions passing automatic validation & 828 \\
Questions requiring manual adjudication & 2 \\
\bottomrule
\end{tabular*}
\end{table}

Table~\ref{tab:app_qrel_stats} summarizes the resulting reference. Automatic
validation checks document identity, quoted text, character offsets, slot
coverage, and substring consistency; 828 of 830 questions pass automatically,
with two manually adjudicated under the same criterion.

A slot reaches Locate when any accepted fragment aligned to that slot appears
in a model-visible observation from the corresponding original document. Both
reference fragments and observed text are normalized using Unicode NFKC
normalization, case folding, whitespace collapse, and typographic quote/dash
normalization, followed by exact normalized substring matching within the same
document. No embedding model or LLM judge is used for matching.

This reference is an operational benchmark target rather than an exhaustive
semantic oracle: semantically equivalent evidence paths not included in the
frozen annotation may fall outside the target. Final-answer accuracy is
therefore reported independently of Locate.

\subsection{Building the Persistent Multi-View Corpus Atlas}
\label{app:atlas_signatures}
\label{app:atlas_construction}

Atlas construction is performed once per corpus using only original document
identifiers, URLs, and text; evaluation questions, answers, Qrels,
trajectories, and correctness labels are excluded. For each document, AtlasNav
constructs Topic, Identity, Episode, and Relation signatures from cleaned and
deduplicated passages. View-specific passage selection is supplemented with
uniformly sampled backfill passages so that lexical selection does not become
a hard information boundary.

Each signature is embedded with Qwen3.7-Text-Embedding
\citep{alibabacloud2026qwen37embedding} and projected with a view-specific PCA
transform. The resulting representations independently induce sparse HNSW
neighborhood graphs \citep{malkov2020hnsw}. AtlasNav then applies multiplex
Leiden clustering: Topic and Identity define parent regions, while Episode and
Relation refine each parent into leaf regions. The frozen BrowseComp-Plus
construction settings are summarized in Table~\ref{tab:app_atlas_config}.

\begin{table}[!htbp]
\caption{Frozen Atlas construction configuration on BrowseComp-Plus.}
\label{tab:app_atlas_config}
\centering
\small
\begin{tabular*}{\textwidth}{@{\extracolsep{\fill}}lp{0.70\textwidth}@{}}
\toprule
Component & Setting \\
\midrule
View signatures & Topic (6,144 characters); Identity, Episode, Relation
                  (4,096 characters each) \\
Encoding & 2,560-dimensional unit vectors; independent 192-dimensional PCA
           projections fitted on 50,000 deterministic samples \\
Sparse graphs & cosine/IP HNSW; $k=48$, $M=32$,
                \texttt{efConstruction}$=160$, \texttt{efSearch}$=128$ \\
Coarse hierarchy & multiplex Topic (1.0) + Identity (0.75); 77 parent regions \\
Fine hierarchy & conditional Episode + Relation within each parent; 443 leaves \\
Addressing & 100,195/100,195 documents receive one parent--leaf address;
             canonical files remain directly searchable and openable \\
\bottomrule
\end{tabular*}
\end{table}

Every document receives a stable parent--leaf address. Each region stores a
stable identifier, representative anchors, and cross-region bridges. Anchors
combine graph centrality, proximity to multi-view region centroids, and an
MMR-style diversity penalty \citep{carbonell1998mmr}; bridges preserve strong
Episode- and Relation-level connections across leaves. The Atlas adds this
reusable navigation structure without pruning the underlying corpus, and
original documents remain directly searchable and openable during interaction.

\subsection{Query-Adaptive Routing}
\label{app:router_tasks}
\label{app:routing}

The persistent Atlas defines the available navigation directions, while the
query-adaptive router determines how strongly each direction should influence
the current query. As described in Section~\ref{sec:atlasnav_navigation}, the
router outputs weights for Topic, Identity, Episode, Relation, and BM25, which
are combined through the weighted RRF in Eq.~(\ref{eq:weighted_rrf}).

\paragraph{Corpus-derived supervision.}
Router supervision is constructed entirely from the frozen corpus and uses no
BrowseComp-Plus evaluation questions, answers, Qrels, trajectories, or
correctness labels. Training tasks require one, two, or three supporting
documents. For multi-document tasks, the positive support set is fixed before
a language model renders the natural-language question, so the rendering step
cannot choose or modify the supervision target. Candidate tasks are filtered
for grounding, answer uniqueness, support necessity, and decoy independence.
Parent regions are split before task construction, yielding parent-disjoint
training, validation, and test sets. Dataset statistics and the frozen router
configuration are summarized in Table~\ref{tab:app_router_data}.

\begin{table}[!htbp]
\caption{Corpus-derived router data and frozen model configuration.}
\label{tab:app_router_data}
\centering
\small
\begin{tabular}{@{}p{0.22\textwidth}p{0.72\textwidth}@{}}
\toprule
Item & Setting \\
\midrule
Tasks & 1,913 single; 4,300 pair; 950 triple; 7,163 total \\
Parent-disjoint split & 4,972 train; 1,167 validation; 1,024 test \\
Input/model & 816 features; 4,902-parameter linear three-head router \\
Output & Topic, Identity, Episode, Relation, and BM25 channel weights \\
Objective & all-positive ranking + worst-positive bottleneck and calibration losses \\
Safety calibration & 0.05 floor on each view's share of semantic mass; low
                     confidence shrinks toward uniform semantic weights and a
                     1:1 semantic/BM25 mass \\
\bottomrule
\end{tabular}
\end{table}

\paragraph{Router parameterization and training.}
The router is a three-head linear model over standardized query and
corpus-derived features. One head allocates weight among the four semantic
views, one controls the semantic-versus-BM25 balance, and one estimates
routing confidence. A minimum semantic-view floor prevents any view from being
fully suppressed, while low-confidence predictions are shrunk toward uniform
semantic allocation and a neutral semantic/BM25 balance. The resulting
nonnegative channel weights are used directly by Eq.~(\ref{eq:weighted_rrf}).

Training combines an all-positive ranking loss, a worst-positive bottleneck
term for multi-support tasks, auxiliary losses for semantic-view allocation,
semantic-versus-lexical mass, and routing confidence, and $L_2$
regularization. The bottleneck term prevents the router from receiving full
credit by retrieving only the easiest member of a multi-document support set.
All hyperparameters are fixed before end-to-end evaluation and shared across
reported settings.

On the 1,024 parent-disjoint test tasks, learned routing improves all-positive
Recall@30 from 0.6465 to 0.6680 relative to the uniform-safe router, with a
larger gain on triple-support tasks (0.3647 to 0.4235). These intrinsic
results are used only to verify the routing component; the paper's end-to-end
claims rely on the controlled interface comparisons in
Section~\ref{sec:experiments}.

\subsection{Online Atlas Navigation}
\label{app:atlas_access_points}
\label{app:online_protocol}

At inference time, the Atlas and router remain frozen. Document priorities
from Eq.~(\ref{eq:weighted_rrf}) are pooled to leaf and then parent regions to
form the initial navigation view, which exposes ten high-priority parent
regions and up to three representative leaf anchors per parent. Its average
observation size is matched to the dynamic-workspace baseline (8,293 vs.\
8,352 bytes). The agent may then expand regions, inspect members, follow
cross-region bridges, reroute, search, and open original documents; full-corpus
lexical search and direct document reads remain available throughout. Online
routing uses only the query and frozen corpus-side statistics and never
accesses gold evidence, Qrels, or correctness labels. The downstream agent
follows the common budget and finalization protocol in
Appendix~\ref{app:experimental_details}.

%% file: sections/appendix_e_main_analysis.tex
\section{Main-Experiment Analysis}
\label{app:complete_results}
\raggedbottom

This appendix complements the three main experiments in
Section~\ref{sec:experiments} with analyses omitted from the main text for
space. The goal is not to introduce additional evaluation settings, but to
examine the reported results from complementary perspectives: budget-level
evidence realization, structural behavior, cost distribution, controlled
scaling, and transfer across heterogeneous enterprise content. Complete
numerical records are provided in Appendix~\ref{app:complete_main_results}.

\subsection{BrowseComp-Plus: Evidence Dynamics and Structural Diagnostics}

\subsubsection{Budget-Level Evidence Realization}
\label{app:budget_dynamics}

\begin{figure}[H]
\centering
\includegraphics[width=.93\textwidth]{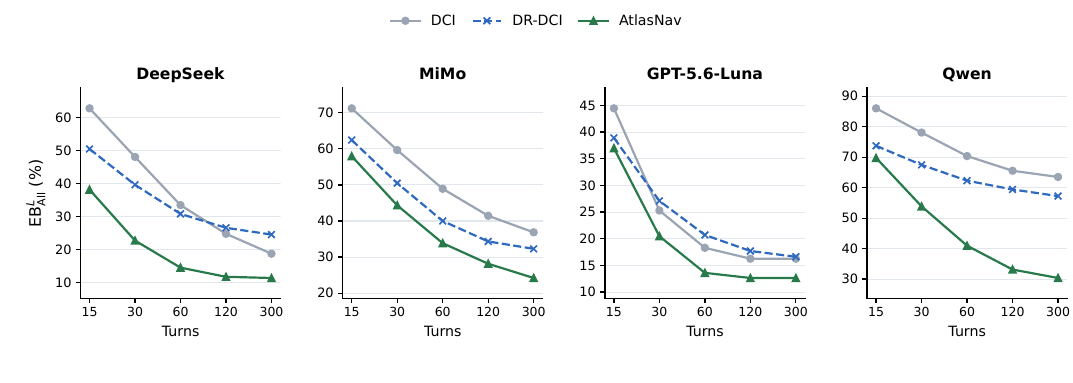}
\caption{BrowseComp-Plus Locate Evidence Blindness
$\operatorname{EB}_{\mathrm{All}}^L$ over interaction turns for DCI, DR-DCI,
and AtlasNav; lower is better. Exact checkpoint values are reported in
Appendix~\ref{app:complete_main_results}.}
\label{fig:app_locate_dynamics}
\end{figure}

Figure~\ref{fig:app_locate_dynamics} extends the finite-budget analysis to
complete Locate Evidence Blindness across all four backbones. AtlasNav reaches
a lower $\operatorname{EB}_{\mathrm{All}}^L$ earlier and generally preserves
this advantage as interaction proceeds. On DeepSeek, for example, Locate
blindness is 38.19\% at 15 turns and 22.77\% at 30 turns, compared with
50.48\% and 39.64\% for DR-DCI. The same qualitative pattern holds for MiMo,
GPT-5.6-Luna, and Qwen; exact checkpoint values are reported in
Appendix~\ref{app:complete_main_results}.

\begin{table}[H]
\caption{Paired uncertainty for the AtlasNav--DR-DCI endpoint comparison on
the two backbones with retained per-question records. Confidence intervals are
fixed-seed paired bootstrap intervals; $p$-values are from exact McNemar tests.}
\label{tab:app_endpoint_uncertainty}
\centering
\small
\begin{tabular*}{\textwidth}{@{\extracolsep{\fill}}lrrrrr@{}}
\toprule
Backbone & \shortstack{AtlasNav\\Acc.} & \shortstack{DR-DCI\\Acc.} & Difference
& \shortstack{Paired bootstrap\\95\% CI} & Exact McNemar \\
\midrule
DeepSeek & 92.05\% & 84.58\% & $+7.47$ pp & $[4.94,10.00]$ pp
         & $p=1.22\times10^{-8}$ \\
GPT-5.6-Luna  & 91.81\% & 87.83\% & $+3.98$ pp & $[1.69,6.39]$ pp
         & $p=0.00119$ \\
\bottomrule
\end{tabular*}
\end{table}

Table~\ref{tab:app_endpoint_uncertainty} quantifies endpoint uncertainty for
the two backbones with retained paired per-question records. AtlasNav improves
strict accuracy over DR-DCI by 7.47 points on DeepSeek, with a paired bootstrap
95\% confidence interval of $[4.94,10.00]$ points and exact McNemar
$p=1.22\times10^{-8}$. On GPT-5.6-Luna, the improvement is 3.98 points, with a
95\% interval of $[1.69,6.39]$ and $p=0.00119$. We do not reconstruct
significance tests for MiMo or Qwen because complete paired instance-level
records are unavailable.

\begin{figure}[H]
\centering
\includegraphics[width=.93\textwidth]{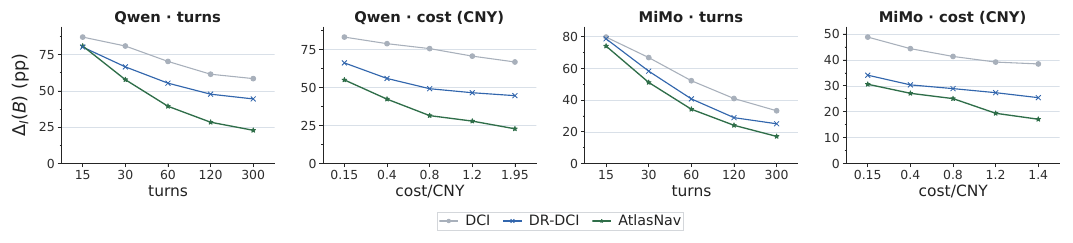}
\caption{Finite-budget empirical reference gaps $\Delta_I(B)$ for Qwen and MiMo
under turn and query-time inference-cost budgets. These panels complement
Figure~\ref{fig:budget_convergence} in the main text; lower is better.}
\label{fig:app_budget_gap}
\end{figure}

To complement the GPT-5.6-Luna and DeepSeek reference-gap curves in the main
text, Figure~\ref{fig:app_budget_gap} reports the corresponding Qwen and MiMo
trajectories under turn and query-time inference-cost budgets. Both backbones
show the same earlier convergence pattern: AtlasNav closes the empirical
reference gap faster than the compared interfaces under finite interaction.

\subsubsection{Structural Diagnostics}
\label{app:ablations}

\begin{table}[H]
\caption{Hierarchical view-assignment ablation on a shared 166-question
subset. All variants use the same corpus, view graphs, router, agent, and
observation budget and differ only in parent--child view assignment.}
\label{tab:app_hierarchy_assignment}
\label{app:hierarchy_assignment}
\centering
\small
\begin{tabular*}{.86\textwidth}{@{\extracolsep{\fill}}llrr@{}}
\toprule
Parent views & Child views & Accuracy $\uparrow$ & Mean turns $\downarrow$ \\
\midrule
\textbf{Topic + Identity} & \textbf{Episode + Relation} & \textbf{95.78} & \textbf{27.43} \\
Topic + Episode & Identity + Relation & 89.16 & 37.33 \\
Topic + Relation & Identity + Episode & 91.57 & 35.10 \\
Identity + Episode & Topic + Relation & 91.57 & 35.88 \\
Identity + Relation & Topic + Episode & 89.76 & 31.89 \\
Episode + Relation & Topic + Identity & 88.55 & 36.02 \\
\bottomrule
\end{tabular*}
\end{table}

Table~\ref{tab:app_hierarchy_assignment} tests whether the parent--child
assignment of the four semantic views matters. Topic+Identity /
Episode+Relation achieves the highest accuracy (95.78\%) and the fewest mean
turns (27.43) among the six assignments. Compared with the exact role reversal,
Episode+Relation / Topic+Identity, the selected hierarchy answers 159 versus
147 of 166 questions correctly, with exact McNemar $p=0.00754$. This supports
the selected hierarchy on BrowseComp-Plus without implying that the same
ordering is universally optimal.

\begin{table}[H]
\caption{Mean number of distinct Atlas parent (P) and leaf (L) regions under
matched surfaced-file budgets on BrowseComp-Plus.}
\label{tab:app_region_diversity}
\centering
\small
\begin{tabular*}{.92\textwidth}{@{\extracolsep{\fill}}lrrrrrr@{}}
\toprule
Interface & P@10 & L@10 & P@20 & L@20 & P@50 & L@50 \\
\midrule
DCI & 7.77 & 8.64 & 12.96 & 15.77 & 22.83 & \textbf{33.84} \\
DR-DCI & 3.95 & 5.50 & 5.97 & 8.85 & 9.89 & 15.80 \\
\textbf{AtlasNav} & \textbf{9.00} & \textbf{9.99} & \textbf{16.54} & \textbf{18.64} & \textbf{23.53} & 31.31 \\
\bottomrule
\end{tabular*}
\end{table}

Table~\ref{tab:app_region_diversity} examines how broadly each interface
exposes Atlas regions early in the trajectory. AtlasNav covers the most
distinct parent and leaf regions among the first 10 and 20 surfaced files,
while the gap narrows by 50 files and DCI slightly exceeds it in leaf
diversity. The benefit therefore lies primarily in broadening early
exploration, when limited observations are most consequential, rather than
maximizing region diversity over long trajectories.

\subsubsection{Cost Distribution and Reusable Preparation}
\label{app:offline_resources}

\begin{table}[!htbp]
\caption{Per-question query-time agent inference cost distributions and
reusable embedding-cost amortization. Panel (a) excludes judge cost, offline
preparation, and query embedding; panel (b) includes required query embedding
in its operating-cost values.}
\label{tab:app_cost_reuse}
\centering
\small
\textbf{(a) Per-question agent-cost distributions on aligned backbones.}\\[3pt]
\begin{tabular*}{\textwidth}{@{\extracolsep{\fill}}llrrrr@{}}
\toprule
Backbone & Interface & Mean / query & Median / query & Correct mean & Incorrect mean \\
\midrule
DeepSeek (CNY) & Raw DCI & 1.479757 & 0.362830 & 0.873155 & 4.142503 \\
& DR-DCI & 0.903333 & 0.194745 & 0.568829 & 2.737881 \\
& AtlasNav & 0.630375 & 0.187247 & 0.464648 & 2.548784 \\
\midrule
GPT-5.6-Luna (USD) & Raw DCI & 0.159476 & 0.030587 & 0.109962 & 0.501356 \\
& DR-DCI & 0.108866 & 0.014706 & 0.055269 & 0.495726 \\
& AtlasNav & 0.092821 & 0.015923 & 0.057562 & 0.487934 \\
\bottomrule
\end{tabular*}

\vspace{7pt}
\textbf{(b) BrowseComp-Plus reusable embedding-cost amortization under DeepSeek.}\\[3pt]
\begin{tabular*}{.88\textwidth}{@{\extracolsep{\fill}}lrrr@{}}
\toprule
Quantity & AtlasNav & DR-DCI & Raw DCI \\
\midrule
Operating cost / query & CNY 0.6307 & CNY 0.9034 & CNY 1.4798 \\
Reusable embedding subtotal & CNY 127.68 & -- & -- \\
AtlasNav saving / query & -- & CNY 0.2727 & CNY 0.8490 \\
Gross break-even queries & -- & \textbf{469} & \textbf{151} \\
\bottomrule
\end{tabular*}

\vspace{3pt}
\begin{minipage}{.94\textwidth}
\footnotesize
The AtlasNav reusable subtotal comprises CNY 126.78 for four-view corpus
encoding and CNY 0.90 for router-supervision query encoding. It is not a
complete offline construction bill.
\end{minipage}
\end{table}

Table~\ref{tab:app_cost_reuse} shows that AtlasNav reduces query-time agent
inference cost relative to DR-DCI on both aligned backbones, but the
distributional pattern differs. On DeepSeek, both mean and median cost
decrease, whereas on GPT-5.6-Luna the mean reduction is driven mainly by fewer
expensive long-tail trajectories. Thus, the aggregate saving should not be
interpreted as every query becoming uniformly cheaper.

AtlasNav also incurs reusable corpus-side preparation. We report this
separately from query-time agent inference cost because the two have different
scaling behavior: operating cost is incurred for every query, whereas Atlas
construction is performed once and reused across subsequent queries over the
same corpus. Treating the latter as a per-query online cost would therefore
depend arbitrarily on how many queries happen to share the representation. To
make this trade-off explicit rather than omit it,
Table~\ref{tab:app_cost_reuse} reports an amortization analysis using the
reconstructable preparation costs. The reusable embedding subtotal is
CNY~127.68; after including required per-query embedding, gross break-even
occurs after 469 reused queries relative to DR-DCI and 151 relative to Raw
DCI. These values characterize the recorded pipeline and should not be
interpreted as complete lifecycle-cost accounting.

\subsubsection{Evidence Realization and Final Correctness}
\label{app:evaluation_boundaries}

\begin{table}[!htbp]
\caption{Representative BrowseComp-Plus trajectories illustrating the
distinction between evidence realization and final-answer correctness.}
\label{tab:app_cases}
\centering
\small
\begin{tabular}{@{}p{0.12\textwidth}p{0.34\textwidth}p{0.17\textwidth}p{0.28\textwidth}@{}}
\toprule
Case & Evidence state & Outcome & Diagnostic \\
\midrule
8 (\emph{Lush Life}) & AtlasNav and DCI realize the complete annotated
evidence; DR-DCI does not realize the annotated chain. & AtlasNav correct;
DCI and DR-DCI wrong. & Evidence access and downstream reasoning are distinct
failure modes. \\
394 (wing three-quarter) & AtlasNav reaches only part of the annotated
supporting-document set but realizes the complete decision-relevant slot-level
evidence; the baselines do not. & AtlasNav correct; baselines wrong. & Complete
support-document coverage is not required once the decisive evidence slots
are realized. \\
417 (Owlman) & AtlasNav realizes the complete annotated evidence; DR-DCI
realizes only part of it. & AtlasNav wrong; DR-DCI correct. & Complete
annotated evidence does not guarantee correct synthesis, and partial annotated
realization does not preclude a correct answer. \\
\bottomrule
\end{tabular}
\end{table}

Table~\ref{tab:app_cases} illustrates why Evidence Blindness and final-answer
accuracy should be interpreted jointly rather than interchangeably. The cases
show that complete evidence realization does not guarantee correct synthesis,
while incomplete realization of the frozen benchmark reference does not
necessarily preclude a correct answer. Evidence Blindness therefore diagnoses
whether benchmark-grounded evidence becomes model-visible during interaction,
whereas final accuracy measures whether the agent ultimately reasons from the
available information successfully.

\FloatBarrier
\subsection{PhantomWiki: Where Scaling Breaks the Funnel}

PhantomWiki isolates corpus growth while keeping the 200 questions and their
supporting documents fixed. As shown in
Section~\ref{sec:phantomwiki_results}, AtlasNav retains the lowest Surface
Evidence Blindness and the highest strict accuracy across all evaluated scales.
The supplementary Open results in Appendix~\ref{app:complete_main_results}
show where degradation emerges at the largest scale: at roughly one million
files, AtlasNav has 49.5\% Surface blindness but 71.0\% Open blindness, and
only 57.4\% of questions with complete Surface proceed to complete Open. Thus,
persistent organization continues to expose relevant directions, while the
finite interaction budget increasingly limits how many surfaced candidates can
actually be opened. Because PhantomWiki lacks fragment-level annotations
comparable to the BrowseComp-Plus reference, we restrict this analysis to
Surface and Open rather than define a separate Locate criterion.

\FloatBarrier
\subsection{EnterpriseRAG-Bench: Transfer Across Question Types and Sources}

\begin{table}[!htbp]
\caption{EnterpriseRAG-Bench Overall score by question category and
overlapping source tag. Source tags are not mutually exclusive; the full-set
Overall score is 87.00. Complete metric matrices are reported in
Appendix~\ref{app:complete_main_results}.}
\label{tab:app_enterprise_landscape}
\centering
\small
\begin{tabular*}{\textwidth}{@{\extracolsep{\fill}}lrlr@{}}
\toprule
Question category & Overall & Source tag & Overall \\
\midrule
Miscellaneous & 96.08 & HubSpot & 92.42 \\
Intra-document & 96.46 & Gmail & 86.74 \\
Constrained & 94.57 & Jira & 81.21 \\
Basic & 95.16 & Google Drive & 88.70 \\
Conflicting information & 92.69 & Fireflies & 87.22 \\
Information not found & 100.00 & Linear & 84.23 \\
Semantic & 80.73 & GitHub & 82.80 \\
Completeness & 66.98 & Slack & 80.03 \\
Project-related & 59.09 & Confluence & 78.00 \\
High-level & 58.00 & & \\
\bottomrule
\end{tabular*}
\end{table}

Table~\ref{tab:app_enterprise_landscape} decomposes EnterpriseRAG-Bench
performance by question category and overlapping source tag. The category
results reveal a clear boundary: AtlasNav remains strong on miscellaneous,
intra-document, constrained, basic, and conflicting-information questions, but
performs substantially worse on project-related, completeness-oriented, and
high-level questions. These harder categories generally require broader
aggregation across distributed evidence rather than locating a small number of
decisive items.

Source-tag variation is comparatively smaller. Overall scores range from 78.00
on Confluence-tagged questions to 92.42 on HubSpot-tagged questions, with no
source family exhibiting a drop comparable to the hardest question categories.
Because source tags overlap and their sample sizes differ, these slices are
descriptive rather than independent test sets. Taken together, the results
suggest that the remaining difficulty is associated more with
evidence-aggregation demands than with any particular enterprise source or
connector.

\FloatBarrier
\flushbottom

%% file: sections/appendix_f_boundary.tex
\section{Broader Positioning and Boundary Regimes}
\label{app:auxiliary}
\raggedbottom
\setlength{\intextsep}{4pt plus 1pt minus 1pt}
\setlength{\abovecaptionskip}{2pt}
\setlength{\belowcaptionskip}{2pt}

This appendix places AtlasNav within the broader corpus-access design space and
examines three settings in which the dominant bottleneck differs from
BrowseComp-Plus. The goal is to clarify both the methodological position of
persistent navigation and the regimes in which its benefits change form.

\subsection{Positioning and Controlled Comparison Axis}

Table~\ref{tab:app_access_design} summarizes the surrounding corpus-access
design space. Retrieval and reranking methods primarily select evidence for a
bounded context; structured RAG adds reusable corpus organization but still uses
that structure mainly for retrieval-mediated selection. Interactive methods
retain continued corpus access after partial evidence is observed. Within this
space, AtlasNav combines persistent corpus-side structure with continued
interaction: the representation is built once, while each query changes only
which parts of that structure are prioritized during navigation.

\begin{table}[H]
\caption{Conceptual corpus-access design space. The taxonomy positions methods
by interface semantics rather than comparing performance across heterogeneous
experimental stacks.}
\label{tab:app_access_design}
\centering
\scriptsize
\setlength{\tabcolsep}{2.0pt}
\renewcommand{\arraystretch}{1.12}
\begin{tabular}{@{}p{.78in}p{1.18in}p{.62in}p{.62in}p{.68in}p{1.05in}@{}}
\toprule
Paradigm & Representative methods & \shortstack{Persistent\\structure} &
\shortstack{Query-specific\\structure} & \shortstack{Continued\\interaction} &
Primary role of structure \\
\midrule
\shortstack[l]{Sparse/dense\\retrieval} & BM25; dense retrievers & Index only & No & Limited &
Select top-$k$ evidence \\
\shortstack[l]{Retrieval +\\reranking} & ReasonRank \citep{liu2026reasonrank}; neural rerankers & Index & No & Limited &
Refine evidence ranking \\
\shortstack[l]{Hierarchical/\\graph RAG} & RAPTOR; HippoRAG; SiReRAG; LinearRAG \citep{zhuang2026linearrag} & Yes & Usually no &
Retrieval-mediated & Structured evidence selection \\
\shortstack[l]{Trained retrieval\\agents} & \shortstack[l]{Search-R1; R1-Searcher \citep{song2025r1searcher};\\ASearcher \citep{gao2025asearcher}} & Retrieval index &
Adaptive querying & Via search API & Improve search policy \\
\shortstack[l]{Relevance-guided\\interaction} & RARG \citep{li2026rarg} & Retrieval index &
Query-conditioned priority & Yes & Guide traversal and local match visibility \\
\shortstack[l]{Raw direct\\interaction} & DCI & No & No & Yes & Direct corpus operations \\
Dynamic workspace & DR-DCI & Shared index & Yes & Yes & Query-time local organization \\
\shortstack[l]{Persistent\\navigation} & \textbf{AtlasNav} & Yes & Priority only & Yes & Reusable interactive organization \\
\bottomrule
\end{tabular}
\end{table}

The main controlled comparison therefore begins after continued corpus
interaction is already available. DCI provides raw interactive access with
little reusable organization, DR-DCI adds query-conditioned organization through
dynamic workspace construction, and AtlasNav moves reusable organization before
query time while preserving continued access to the original corpus. Holding the
downstream backbone, corpus, interaction budget, and evaluation protocol fixed
makes this progression isolate the effect of when and how corpus structure is
introduced. Structured RAG and retrieval-agent methods remain relevant points in
the broader design space, but their interfaces, readers, retrieval budgets, and
context-construction procedures differ too substantially for published scores to
serve as controlled head-to-head baselines here.

\subsection{2WikiMultiHopQA: High-Recall Multi-Hop Reasoning}

2Wiki-Global tests a regime in which evidence access is already strong. We merge
the local contexts of 400 questions \citep{ho2020constructing} into a shared
56,684-document corpus so that DCI, DR-DCI, and AtlasNav search the same global
collection. Endpoint accuracy is nearly saturated: DCI reaches 95.50\% strict
accuracy, while DR-DCI and AtlasNav both reach 94.50\%. The main difference is
interaction cost. AtlasNav uses CNY~10.89 and 3,135 total turns, compared with
CNY~22.59 and 4,117 turns for DCI and CNY~28.01 and 5,540 turns for DR-DCI.

\begin{table}[H]
\caption{Exact 2Wiki-Global results on 400 questions over a shared
56,684-document corpus. Cost is total recorded query-time agent inference cost
in CNY.}
\label{tab:app_2wiki}
\centering
\small
\begin{tabular*}{.72\textwidth}{@{\extracolsep{\fill}}lrrr@{}}
\toprule
Interface & Strict Acc. $\uparrow$ & Cost $\downarrow$ & Turns $\downarrow$ \\
\midrule
DCI & \textbf{95.50} & 22.59 & 4,117 \\
DR-DCI & 94.50 & 28.01 & 5,540 \\
\textbf{AtlasNav} & 94.50 & \textbf{10.89} & \textbf{3,135} \\
\bottomrule
\end{tabular*}
\end{table}

This regime shows that persistent navigation need not improve final accuracy
when evidence access is already close to saturation. Its benefit instead appears
as reduced interaction required to reach a similarly strong evidence state. The
result therefore bounds the main claim: when Evidence Blindness is weak,
persistent organization primarily improves efficiency rather than endpoint
performance.

\subsection{FanOutQA: Many-Answer Aggregation}

FanOutQA shifts the bottleneck from locating a small decisive evidence chain to
collecting and consolidating many answer items \citep{zhu2024fanoutqa}. Over the
310-question, 1,594-document corpus, DR-DCI achieves the highest Loose and
Strict coverage, at 81.42 and 44.19, while AtlasNav reaches 80.10 and 40.32 with
substantially fewer turns (4,218 versus 6,493 for DR-DCI and 6,596 for DCI).
AtlasNav also does not reproduce its BrowseComp-Plus Surface advantage: complete
Surface/Open blindness is 12.90\%/33.87\%, compared with 6.45\%/28.71\% for
DR-DCI.

\begin{table}[H]
\caption{Exact FanOutQA endpoint results on 310 questions. Loose and Strict are
the official answer-coverage metrics; cost is recorded query-time agent
inference cost in CNY.}
\label{tab:app_fanout}
\centering
\small
\begin{tabular*}{.84\textwidth}{@{\extracolsep{\fill}}lrrrr@{}}
\toprule
Interface & Loose $\uparrow$ & Strict $\uparrow$ & Cost $\downarrow$ & Turns $\downarrow$ \\
\midrule
DCI & 79.65 & 39.35 & 73.35 & 6,596 \\
DR-DCI & \textbf{81.42} & \textbf{44.19} & 43.24 & 6,493 \\
\textbf{AtlasNav} & 80.10 & 40.32 & \textbf{36.49} & \textbf{4,218} \\
\bottomrule
\end{tabular*}
\end{table}

The limiting factor in this regime is therefore not primarily evidence
discovery. Candidate evidence is relatively easy to expose in the compact
corpus, but success requires repeatedly consuming, retaining, and aggregating
many items before finalization. Persistent navigation reduces traversal effort,
yet it does not remove the downstream aggregation bottleneck, so lower
interaction does not translate into higher endpoint coverage.

\subsection{TREC-COVID: Graded Retrieval}

TREC-COVID changes the objective from answer generation to graded document
ranking over 171,332 scientific records \citep{roberts2020treccovid}. We
evaluate nDCG@10 using the official graded relevance judgments.

\begin{figure}[H]
\centering
\includegraphics[width=\textwidth]{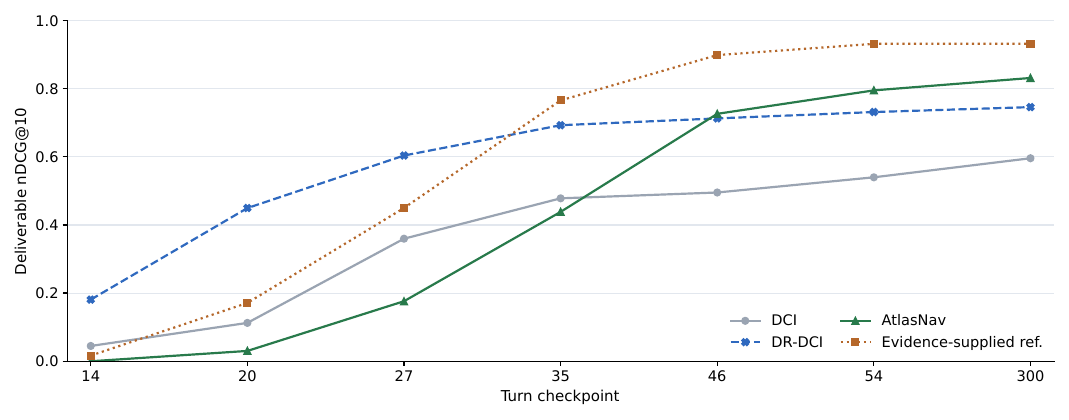}
\caption{Deliverable nDCG@10 over frozen, data-driven turn checkpoints on
TREC-COVID. Unfinished queries score zero at each checkpoint. The
evidence-supplied reference receives the positive-document pool but not graded
labels or the ideal ranking.}
\label{fig:app_trec_turns}
\end{figure}

Figure~\ref{fig:app_trec_turns} shows a delayed crossover: DR-DCI performs
better at the earliest checkpoints, while AtlasNav overtakes it at turn 46 and
reaches 0.8314 nDCG@10 at the endpoint, compared with 0.7459 for DR-DCI and
0.5958 for DCI. The endpoint gain over DR-DCI is 0.0855, with a 95\% confidence
interval of $[0.0140,0.1684]$.

This delayed advantage reflects a different evidence-access regime. Graded
ranking requires allocating interaction across a broader set of relevant
documents and preserving their relative quality, rather than locating a single
sufficient evidence chain. Persistent organization therefore becomes more
useful later, after the agent must move beyond the first locally relevant region
and assemble a stronger ranked set. AtlasNav does not reduce query-time
inference cost relative to DR-DCI in this setting, so TREC-COVID supports the
representation and evidence-allocation benefit of persistent navigation rather
than a universal efficiency advantage.

Together, these boundary regimes show that the effect of persistent navigation
depends on where the dominant bottleneck lies. When evidence discovery is
already strong, as in 2Wiki-Global, the main benefit is reduced interaction;
when exhaustive aggregation dominates, as in FanOutQA, navigation alone does not
improve endpoint coverage; and when broad graded evidence allocation is
required, as in TREC-COVID, the benefit emerges later in the trajectory. The
strongest joint gains in accuracy and efficiency therefore occur when
finite-budget evidence discovery itself remains a major bottleneck, as on
BrowseComp-Plus.

\FloatBarrier
\flushbottom

%% file: sections/appendix_g_prompts.tex
\section{Notation and Prompts}
\label{app:prompts}

We first summarize the principal notation used in Sections~3--4, then include
the full prompts used for the reported evaluations. The system
prompt is issued to the AtlasNav agent at the start of each trajectory, and the
finalization prompt is issued once the budget trigger defined in
Appendix~\ref{app:finalization_protocol} is reached.

\subsection{Notation Summary}
\label{app:notation_summary}

Table~\ref{tab:notation_summary} collects the symbols that readers need to
track across the Evidence Blindness formalization, Atlas construction, and
query-adaptive routing. Checkpoint superscripts $C,S,O,L$ denote Construction,
Surface, Open, and Locate, respectively.

\begin{table}[H]
  \caption{Principal notation used in Sections~3--4.}
  \label{tab:notation_summary}
  \centering
  \small
  \setlength{\tabcolsep}{3pt}
  \renewcommand{\arraystretch}{1.08}
  \begin{tabular*}{\textwidth}{@{\extracolsep{\fill}}p{0.16\textwidth}p{0.31\textwidth}p{0.16\textwidth}p{0.31\textwidth}@{}}
    \toprule
    Symbol & Meaning & Symbol & Meaning \\
    \midrule
    $q$ & Question or query. & $\mathcal{Q}$ & Set of evaluation questions. \\
    $\mathcal{E}_q=\{e_1,\ldots,e_{m_q}\}$ & Required evidence slots for $q$; $m_q$ is their number. & $\tau\in\mathcal{T}=\{C,S,O,L\}$ & Construction, Surface, Open, or Locate checkpoint. \\
    $\mathcal{E}_q^\tau$ & Evidence slots realized by checkpoint $\tau$. & $r_q^\tau$ & Fraction of required slots realized at $\tau$. \\
    $\operatorname{EB}_{\mathrm{Any/Mean/All}}^\tau$ & Evidence Blindness metrics for total failure, mean loss, and incomplete realization. & $\mathcal{C}$ & Original accessible corpus. \\
    $M$ & Language-model agent. & $I$ & Corpus interface. \\
    $B$ & Interaction budget. & $A_I(B)$ & Strict task performance of $I$ under budget $B$. \\
    $A_{\mathrm{ref}}$ & Same-agent evidence-supplied empirical reference performance. & $\Delta_I(B)$ & Empirical reference gap, $A_{\mathrm{ref}}-A_I(B)$. \\
    $\mathcal{D}$ & Set of original documents. & $\mathcal{V}$ & Four semantic views: $\mathrm{Top}$, $\mathrm{Id}$, $\mathrm{Epi}$, and $\mathrm{Rel}$. \\
    $v\in\mathcal{V}$ & Semantic-view index. & $\operatorname{Sig}_v(d)$ & View-specific signature of document $d$. \\
    $z_d^v$ & View-specific representation of $d$. & $G^v=(\mathcal{D},E^v,W^v)$ & View-$v$ neighborhood graph, with edges $E^v$ and affinities $W^v$. \\
    $\mathcal{P}$ & Set of coarse parent regions. & $\mathcal{L}(P)$ & Leaf regions nested within parent $P$. \\
    $G^v[P]$ & View-$v$ subgraph restricted to $P$. & $c\in\mathcal{V}\cup\{\mathrm{BM25}\}$ & Routing channel index. \\
    $\operatorname{rank}_c(d\mid q)$ & Rank of $d$ for $q$ under channel $c$. & $w_c(q)$ & Query-dependent router weight for $c$. \\
    $s_q(d)$ & Fused query-adaptive priority for $d$. & $\kappa$ & Reciprocal Rank Fusion smoothing constant. \\
    \bottomrule
  \end{tabular*}
\end{table}

\clearpage
\subsection{AtlasNav System Prompt}
\label{app:prompt_system}

\begin{quote}
You are solving a closed-corpus research question. The Atlas is a complete map,
not a candidate list: every canonical file remains reachable through region
pagination and full-text search. Region names, filenames, rankings, and previews
are navigation hints, never final evidence. Open canonical documents and verify
the decisive fields before answering. Prefer complementary regions and evidence
chains over many near-duplicate high-ranked files. Keep research focused and
submit the shortest sufficient answer as
\mbox{\textless{}FINAL\textgreater{}answer\textless{}/FINAL\textgreater{}}.
\end{quote}

\subsection{Budget-Reached Finalization Prompt}
\label{app:prompt_finalization}

\begin{quote}
The preregistered research budget has been reached. Safely release the task now:
do not start new discovery. Review the evidence already visible, resolve the
answer as well as possible, and return exactly one concise
\mbox{\textless{}FINAL\textgreater{}answer\textless{}/FINAL\textgreater{}}. Do
not discuss the budget protocol.
\end{quote}